\documentclass[letterpaper,journal]{IEEEtran}
\usepackage{amsmath,amsfonts}
\usepackage{array}
\usepackage[caption=false,font=normalsize,labelfont=sf,textfont=sf]{subfig}
\usepackage{textcomp}
\usepackage{stfloats}
\usepackage{url}
\usepackage{verbatim}
\usepackage{graphicx}
\usepackage{cite}
\usepackage{cite}
\usepackage{amsmath,amssymb,amsfonts}
\usepackage{graphicx}
\usepackage{textcomp}
\usepackage{xcolor}
\usepackage{xspace}
\usepackage{enumitem}
\def\BibTeX{{\rm B\kern-.05em{\sc i\kern-.025em b}\kern-.08em
    T\kern-.1667em\lower.7ex\hbox{E}\kern-.125emX}}

\usepackage{xcolor,colortbl}
\usepackage{float}
\usepackage{graphicx}

\usepackage{multirow}
\usepackage{xcolor}
\usepackage{colortbl}
\usepackage{fancybox}

\usepackage{setspace}
\usepackage{enumitem}
\usepackage{color}
\usepackage{diagbox}
\usepackage{soul}
\usepackage{algpseudocode}
\usepackage{framed}
\usepackage{amsmath}
\usepackage{pifont}
\usepackage{tikz}
\usepackage{balance}
\usepackage[normalem]{ulem}
\usepackage{url}
\usepackage[T1]{fontenc}
\usepackage[utf8]{inputenc}
\usepackage[font=small,labelfont=bf,tableposition=top]{caption}
\usepackage{booktabs}
\usepackage{threeparttable}
\usepackage{fontawesome5}
\usepackage{tabularx}
\usepackage[many]{tcolorbox}

\usepackage[ruled,vlined,linesnumbered,longend]{algorithm2e}

\definecolor{grey}{rgb}{0.9,0.9,0.9}
\definecolor{lightgreen}{HTML}{bae4b3}
\definecolor{lightgrey}{HTML}{f0f0f0}
\definecolor{mygreen}{HTML}{31a354}
\definecolor{mygray}{HTML}{666666}

\definecolor{fullgreen}{RGB}{0,160,0}
\definecolor{failred}{RGB}{220,0,0}
\definecolor{partorange}{RGB}{245,120,0}

\newcommand{\cmark}{\textcolor{fullgreen}{\ding{51}}}
\newcommand{\xmark}{\textcolor{failred}{\ding{55}}}

\newcommand{\pmark}{%
\tikz[baseline=-0.55ex]{
  \node[inner sep=0pt, outer sep=0pt] (m) {\textcolor{partorange}{\ding{51}}};
  \draw[partorange, line width=0.8pt, line cap=round]
    ([xshift=0.30em,yshift=0.50em]m.south west) --
    ([xshift=0.50em,yshift=0.20em]m.south west);
}%
}

\definecolor{bestbg}{HTML}{D6EAF8}     
\definecolor{besttext}{HTML}{154360}   
\definecolor{secondbg}{HTML}{EBEDEF}   
\definecolor{secondtext}{HTML}{1C2833} 

\newcommand{\best}[1]{\cellcolor{bestbg}\textcolor{besttext}{\textbf{#1}}}
\newcommand{\second}[1]{\cellcolor{secondbg}\textcolor{secondtext}
{\textbf{#1}}}
\usepackage{tcolorbox}

\makeatletter
\newcommand{\linebreakand}{%
  \end{@IEEEauthorhalign}
  \hfill\mbox{}\par
  \mbox{}\hfill\begin{@IEEEauthorhalign}
}
\makeatother

\newcounter{mysubsubsection}[subsection]
\renewcommand{\themysubsubsection}{\Alph{subsection}.\arabic{mysubsubsection}}

\newcommand{\mysubsubsection}[1]{%
  \refstepcounter{mysubsubsection}%
  \vspace{0.6em}
  \noindent\textit{\textbf{\themysubsubsection.\ #1}}\par
  \vspace{0.2em}
}

\author{Jun~Xiao, Rui~Ye, Dong~Cao, Jinyu~Gu, Ying~Wang\textsuperscript{*}, and Zhiliang~Zhu%
\thanks{\textsuperscript{*}Ying Wang is the corresponding author.}%
\thanks{Jun Xiao, Rui Ye, Dong Cao, Ying Wang, and Zhiliang Zhu are with the Software College, Northeastern University, Shenyang, China. E-mail: xiaojun@mails.neu.edu.cn; yerui@mails.neu.edu.cn; caodong@mails.neu.edu.cn; wangying@swc.neu.edu.cn; ZHUZhiLiang\_NEU@163.com.}%
\thanks{Jinyu Gu is with the School of Computer Science,
Shanghai Jiao Tong University, Shanghai, China,
and also with Shanghai Syslong Information Technology Co., Ltd. E-mail: gujinyu@sjtu.edu.cn.}%
}

\newcounter{stage}
\newcommand{\thestep}{\arabic{stage}}

\usepackage{amsmath,amssymb}

\usepackage[colorlinks]{hyperref}   
 
\definecolor{mygreen}{HTML}{31a354}
\definecolor{revisionpurple}{RGB}{145, 60, 190}

\def\BibTeX{{\rm B\kern-.05em{\sc i\kern-.025em b}\kern-.08em
    T\kern-.1667em\lower.7ex\hbox{E}\kern-.125emX}}

\usepackage{xcolor,colortbl}
\usepackage{float}
\usepackage{graphicx}
\usepackage{multirow}
\usepackage{xcolor}
\usepackage{colortbl}
\usepackage{fancybox}
\hypersetup{colorlinks,linkcolor={blue},citecolor={blue},urlcolor={red}} 
\hypersetup{colorlinks=true, urlcolor=blue}
\hypersetup{colorlinks,linkcolor={blue},citecolor={blue},urlcolor={red}} 
\hypersetup{colorlinks=true, urlcolor=blue}

\AtBeginDocument{
  \hypersetup{pdfborder={0 0 1}, urlcolor=.}  
}

\begin{document}

\title{\textsc{VLA-Gauge}: Systematically Evaluating Generalization and Robustness in Vision-Language-Action Models}



\maketitle

\begin{abstract}

Vision-Language-Action models (VLAs) have achieved high success rates on familiar manipulation tasks, yet it remains unclear whether such scores reflect genuine zero-shot generalization and perturbation robustness. Existing evaluation frameworks provide limited capability-level diagnosis and combinatorial perturbation coverage, often without connecting diagnosis to targeted model refinement. We introduce \textsc{VLA-Gauge}, a diagnostic framework that unifies both evaluations under a shared executable task representation. It combines \textcircled{1} paired in-distribution and out-of-distribution tasks across six capability suites, \textcircled{2} single-factor sensitivity analysis with combinatorial perturbation screening and evaluation, and \textcircled{3} diagnosis-driven refinement through targeted data augmentation and model fine-tuning. We evaluate 12 VLAs spanning different model sizes and action-modeling paradigms under a unified protocol. We find that \textcircled{1} The mean Zero-Shot Score on \textsc{VLA-Gauge} is only 22.3\%, with SimpleVLA-RL scoring 8.0\% and ranking second to last among the evaluated models. \textcircled{2} Under three representative combinatorial perturbation configurations, the mean $\boldsymbol{\mathrm{R\text{-}Score}}$ is 9.3. \textcircled{3} Diagnosis-driven fine-tuning improves $\boldsymbol{\pi_{0.5}}$'s zero-shot generalization and robustness under a selected perturbation combination on held-out scenarios by 72.4 and 31.8 scoring points, respectively; the latter exceeds the 10.9-point gain from \textsc{Libero-Plus} single-factor fine-tuning despite fewer successful training trajectories. These results show that high reported success rates on familiar manipulation tasks do not guarantee strong zero-shot generalization to unseen tasks or robustness to combinatorial perturbations, while diagnostic evaluation provides actionable signals for targeted model improvement.

\end{abstract}

\begin{IEEEkeywords}
Benchmark, Perturbation Robustness, Robotic Manipulation, Vision-Language-Action Models, Zero-Shot Generalization.
\end{IEEEkeywords}

\section{Introduction}

\emph{Vision-Language-Action Models} (\textbf{VLAs}) provide a promising path toward building general-purpose robotic policies by jointly modelling \emph{visual observations}, \emph{natural language instructions}, and \emph{robot actions}~\cite{brohan2023rt, driess2023palm, kim2024openvla}. In recent years, with the continued advancement of vision-language pre-training, large-scale robotic datasets, and action generation methods, VLA models have achieved high success rates on familiar manipulation tasks~\cite{fei2026libero-plus, zhou2025libero-pro, wang2026libero-x}. However, a real robotic system must not only complete tasks drawn from familiar distributions, but also generalize to unseen objects, instructions, scenarios, and task compositions, while maintaining stable execution under non-ideal conditions such as visual variation, spatial displacement, and control error. Prior research indicates that changes in lighting, tabletop appearance, and surrounding objects can occur simultaneously in real-world manipulation settings~\cite{pumacay2024colosseum}. Related experiments further show that a model may fail under combinatorial perturbations even when it successfully completes the task under each perturbation in isolation~\cite{sawada2026beyond}. \uline{Therefore, high success rates on familiar manipulation tasks are by themselves insufficient to establish zero-shot generalization to unseen task distributions or execution robustness under combinatorial perturbations, and a systematic evaluation of both aspects is needed.}

Systematic evaluation of VLA models should go beyond aggregate performance on a specific task distribution to separately assess capability-level zero-shot generalization across distributions and execution robustness under perturbations. Recent robotic manipulation benchmarks have expanded beyond standard task completion to evaluate generalization to unseen objects and environments~\cite{li2024evaluating, zhang2025vlabench, zhou2026x-icm} and robustness to variations in appearance, layout, viewpoint, and language~\cite{pumacay2024colosseum, wang2025vlatest, zhang2025vla-arena, wang2026libero-x, zhou2025libero-pro, fei2026libero-plus}. These efforts broaden the scope of evaluation, advance different stages of the diagnostic pipeline, and provide a basis for understanding how models behave under varying task requirements and environmental changes. \uline{It remains an open question, however, how evaluation results can be translated into reliable diagnosis of model weaknesses, targeted improvement measures, and independent verification of whether those improvements take effect.} Viewed against this complete diagnostic objective,  existing evaluations still exhibit the following \emph{\textbf{four limitations}} across different settings: \textcircled{1} 
\emph{\textbf{limited capability attribution:}} task-level success rates aggregate multiple capability demands into a single scalar, rendering performance differences unattributable to specific weaknesses; \textcircled{2} \emph{\textbf{inconsistent distributional evaluation:}} capability dimensions lack unified, paired in-distribution (ID) and out-of-distribution (OOD) tests; \textcircled{3} \emph{\textbf{limited perturbation coverage:}} robustness evaluations consider few perturbation types and parameter settings and remain largely static or single factor; and \textcircled{4} \emph{\textbf{missing diagnosis-to-refinement loop:}} existing benchmarks do not yet integrate fine-grained results from evaluations of model capabilities, ID/OOD performance, and perturbation sensitivity into a closed-loop workflow. Such a workflow guides targeted data augmentation and model optimization based on capability gaps and the conditions under which performance declines, then re-evaluates the model on held-out scenarios.

Addressing these limitations requires a unified diagnostic benchmark for VLAs, whose design poses \emph{\textbf{two core challenges}}: \textcircled{1} \emph{\textbf{Diagnostically attributable task construction:}} Multiple capabilities are coupled in robotic manipulation tasks, and changing predefined factors related to a target capability may also alter the demands on other capabilities. A central challenge is to provide sufficient task diversity within each group and reliably attribute performance differences to specific capability bottlenecks while keeping task goals, success criteria, and non-target requirements as consistent as possible (see~\S\ref{subsec:challenge1}). \textcircled{2} \emph{\textbf{Diagnostic combinatorial perturbation design:}} As the number of perturbation factors and the number of possible values per factor increase, the combination space grows rapidly, making exhaustive enumeration costly. Yet overly aggressive screening may miss important dependencies across perturbation categories. Another central challenge is to select combinations that are affordable to evaluate, interpretable for diagnosis, and representative of real-world deployment conditions while calibrating perturbation strength, ensuring task feasibility, and preserving task semantics (see~\S\ref{subsec:combinatorial_perturbations_design}).

To address these limitations and challenges, we introduce \textsc{VLA-Gauge}, which uses a unified executable task representation to jointly evaluate zero-shot generalization and perturbation robustness for VLA models while supporting diagnosis-driven model refinement. Its \emph{\textbf{key innovations}} include: \textcircled{1} \emph{\textbf{Capability-oriented zero-shot generalization evaluation:}} We design controlled task variants for six capabilities and construct paired ID and OOD tasks. We keep task goals and success criteria as consistent as possible while controlling non-target factors to reduce confounding across capabilities. This design supports evaluation of how well learned capabilities transfer and helps identify generalization bottlenecks. \textcircled{2} \emph{\textbf{Perturbation robustness evaluation:}} We construct a hierarchical perturbation space that satisfies the requirements of task feasibility and semantic consistency, calibrate perturbation strength through single-factor sensitivity analysis, and select representative Perturbation Factor Combinations (PFCs). This makes it possible to examine the joint influence of multiple perturbation categories at an affordable evaluation cost, and provides a basis for identifying potential performance problems under combinatorial perturbations before real deployment.\textcircled{3} \emph{\textbf{Diagnosis-driven model refinement:}} Using the capability gaps found in the earlier fine-grained evaluations of model capability, ID and OOD performance, and perturbation sensitivity, we select challenging scenarios. We then combine these scenarios with the matching optimization suggestions to guide targeted data augmentation and model fine-tuning. Finally, we re-evaluate the model on a held-out set to measure how well these changes work, which closes the loop of diagnosis-driven model optimization.

Using \textsc{VLA-Gauge}, we evaluate the zero-shot generalization and perturbation robustness of 12 VLAs spanning different model sizes and action-modeling paradigms under a unified protocol. Our \emph{\textbf{key findings}} are: \textcircled{1} The mean zero-shot score across all models is only 22.3\%, with 32.9\% on ID tasks and only 7.2\% on OOD tasks. SimpleVLA-RL (e.g., a reported success rate of 99.1\% on \textsc{Libero}) achieves a zero-shot score of only 8.0\% on \textsc{VLA-Gauge}, ranking second to last among the evaluated models. Spatial reasoning is the weakest capability dimension overall, with an average score of 13.0\%, a weakness already evident under ID conditions. Compositional capability performs worst on OOD tasks, with an average score of only 2.8\% and 9 of the 12 models scoring zero. \textcircled{2} Under three representative combinatorial perturbation configurations, the mean $\mathrm{R\text{-}Score}$ is 9.3; 34 out of 36 model-configuration pairs score at or below 17.7, and no model consistently leads across all three configurations. The mean $\mathrm{R\text{-}Score}$ falls from 16.5 to 2.9 as the selected combinations move from one to three perturbation categories. \textcircled{3} Diagnosis-driven fine-tuning improves $\pi_{0.5}$'s zero-shot generalization and robustness under a selected perturbation combination on held-out scenarios by 72.4 and 31.8 scoring points, respectively. The robustness gain substantially exceeds the 10.9-point gain from fine-tuning on single-factor perturbation data from \textsc{Libero-Plus}, despite using fewer successful trajectories. Overall, these results show that high reported in-distribution success rates do not guarantee strong zero-shot generalization to unseen tasks or robustness to combinatorial perturbations, while diagnostic evaluation provides actionable signals for targeted model improvement, validating the need for a unified framework like \textsc{VLA-Gauge}.

In summary, this paper makes the following contributions:

\begin{itemize}[leftmargin=*, itemsep=0.2em, topsep=0.2em]
    \item \textbf{A unified evaluation framework:} We introduce \textsc{VLA-Gauge}, a diagnostic framework for VLA models based on a unified formulation of executable tasks. It integrates capability-oriented paired ID/OOD evaluation, perturbation robustness evaluation, and diagnosis-driven model refinement into a unified closed loop of evaluation, diagnosis, optimization, and re-evaluation.
    \item \textbf{A systematic empirical study:} We systematically evaluate 12 VLA models under a unified protocol, revealing capability-level bottlenecks in zero-shot generalization and robustness to combinatorial perturbations. We validate the effectiveness of diagnosis-driven model refinement through re-evaluation on held-out scenarios and discuss limitations and future research directions for VLA models and their evaluation.
    \item \textbf{Open-source Artifacts:} We release the complete \textsc{VLA-Gauge} evaluation suites, training demonstrations, fine-tuned VLA models, and tools for automated generation and evaluation on our website (\textbf{\url{https://vla-gauge-site.pages.dev/}}). These resources support reproducible evaluation, model optimization, and future research.
\end{itemize}

\section{Related Work}

\begin{table*}[t]
\centering
\caption{\textbf{A systematic comparison of the existing benchmarks.}}
\label{tab:benchmark_comparison}
\vspace{0.15em}

\begin{minipage}{0.96\textwidth}
\centering
\small
\emph{
\textcolor{blue}{\textbf{Focus}}: primary evaluation emphasis;
\textcolor{blue}{\textbf{\#Tasks}}: number of tasks;
\textcolor{blue}{\textbf{\#Objects}}: number of objects;
\textcolor{blue}{\textbf{Fine-Grained}}: fine-grained capability diagnosis;
\textcolor{blue}{\textbf{ID/OOD}}: in-distribution/out-of-distribution zero-shot generalization evaluation;
\textcolor{blue}{\textbf{Robustness}}: robustness testing under diverse perturbations;
\textcolor{blue}{\textbf{Optimization}}: diagnosis-driven model refinement and re-evaluation.
Support level: \cmark: full, \pmark: partial, \xmark: none.
}
\end{minipage}

\vspace{0.25em}

\scriptsize
\setlength{\tabcolsep}{4.2pt}
\renewcommand{\arraystretch}{0.88}

\setlength{\aboverulesep}{0.22ex}
\setlength{\belowrulesep}{0.22ex}

\resizebox{0.94\textwidth}{!}{
\begin{tabular}{llrrcccc}
\toprule
\textbf{Benchmark}
& \textcolor{blue}{\emph{\textbf{Focus}}}
& \textcolor{blue}{\emph{\textbf{\#Tasks}}}
& \textcolor{blue}{\emph{\textbf{\#Objects}}}
& \textcolor{blue}{\emph{\textbf{Fine-Grained}}}
& \textcolor{blue}{\emph{\textbf{ID/OOD}}}
& \textcolor{blue}{\emph{\textbf{Robustness}}}
& \textcolor{blue}{\emph{\textbf{Optimization}}} \\
\midrule
LIBERO~\cite{liu2023libero}        & Knowledge-Transfer  & 130    & 67    & \pmark & \xmark & \xmark & \xmark \\
Behavior-1K~\cite{li2023behavior}   & Long-horizon + Dataset             & 1,000  & 9,331 & \pmark & \xmark & \xmark & \xmark \\
COLOSSEUM~\cite{pumacay2024colosseum}     & Robustness       & 20     & 147    & \pmark & \pmark & \pmark & \pmark \\
SimplerEnv~\cite{li2024evaluating}   & In-distribution Generalization   & 10      & 16    & \pmark & \pmark & \pmark & \xmark \\
VLATest~\cite{wang2025vlatest}     & Robustness   & 4      & 74    & \pmark & \cmark & \pmark & \xmark \\
VLABench~\cite{zhang2025vlabench}      & In-distribution Generalization + Long-horizon & 100    & 2,164 & \pmark & \pmark & \pmark & \xmark \\
VLA-Arena~\cite{zhang2025vla-arena}    & Robustness      & 170    & 122    & \pmark & \pmark & \pmark & \xmark \\
AGNOSTOS~\cite{zhou2026x-icm} & In-distribution Generalization & 23 & 112 & \xmark & \cmark & \xmark & \xmark \\
LIBERO-X~\cite{wang2026libero-x}     & Robustness      & 600    & 96    & \pmark & \pmark & \pmark & \xmark \\
LIBERO-Pro~\cite{zhou2025libero-pro}   & Robustness       & 40     & 77    & \pmark & \cmark & \pmark & \xmark \\
LIBERO-Plus~\cite{fei2026libero-plus}  & Robustness      & 10,030 & 911   & \pmark & \cmark & \pmark & \pmark \\
\midrule
\midrule
\rowcolor{blue!12}
\textbf{VLA-GAUGE}
& \textbf{Zero-shot Generalization + Robustness}
& \textbf{6,300}
& \textbf{3,089}
& \cmark
& \cmark
& \cmark
& \cmark \\

\bottomrule
\end{tabular}
}
\label{comparison}
\vspace{-2.5em}
\end{table*}

\subsection{Generalization and Robustness Challenges of VLA Models}
\label{subsec:baselines}

Recent advances in robot learning have changed the field from task-specific policy networks to general-purpose VLA models~\cite{li2024manipllm,li2024vision}. Pioneering works such as \emph{$\pi$0}~\cite{black2024pi_0} and \emph{OpenVLA}~\cite{kim2024openvla} have contributed to this transition by leveraging large-scale embodied datasets (e.g., \emph{Open-X Embodiment}~\cite{o2024open}) and integrating visual encoders(e.g., \emph{SigLIP}~\cite{zhai2023sigmoid}) with language models. These models map open-vocabulary instructions and visual observations into executable robot actions~\cite{li2026soma}. \uline{However, existing VLA models still differ substantially in their action representations and decoding mechanisms. }They can be categorized into \emph{discrete action models}, \emph{continuous action models}, and \emph{hybrid action models}.

Although these three action modeling paradigms improve the expressiveness of VLA models, they also introduce different generalization and robustness challenges.

\begin{itemize}[leftmargin=*, itemsep=0.2em, topsep=0.2em]
    \item \emph{\textbf{Discrete action models}}:  such as \emph{UniVLA}~\cite{bu2025univla}, \emph{Nora}~\cite{hung2025nora}, \emph{Fast}~\cite{pertsch2025fast}, \emph{DeepThinkVLA}~\cite{yin2025deepthinkvla} and \emph{SimpleVLA-RL}~\cite{li2025simplevla}, quantize continuous control signals into action tokens, which enables them to reuse the autoregressive generation capability of language models. \uline{However, this discretization may discard fine-grained motion information and increase sensitivity to token design and training distributions. This often results in degraded generalization when facing unseen object compositions, complex layouts, or tasks that demand precise control.}
    \item \emph{\textbf{Continuous action models}}: such as \emph{OpenVLA-OFT}~\cite{kim2025fine}, \emph{RIPT-VLA}~\cite{tan2025interactive}, and \emph{CronusVLA}~\cite{li2025cronusvla}, directly generate low-level control signals or action fragments, making them closer to real robotic control interfaces. \uline{However, they are more susceptible to visual observation shifts, variations in initial states, and accumulated execution errors, which can reduce control stability under out-of-distribution scenarios and perturbed conditions.}
    \item \emph{\textbf{Hybrid-action models}}: such as \emph{$\pi$0.5}~\cite{intelligence2025pi_}, combine high-level semantic decision-making with low-level continuous control. \uline{However, this combination may amplify inconsistencies between semantic grounding, target selection, and action execution, making failure sources more coupled and difficult to locate.}
\end{itemize}

Although the three types of models differ in action representation and decoding mechanisms, they all reveal an unstable mapping among vision-language understanding, action representation, and control execution in VLA models. 
\uline{This indicates that the generalization challenge of VLA models arises from both unseen tasks and their dependence on training data compositions (objects, scenarios, and actions), while their robustness challenge extends across perception, decision-making, and execution, rather than being confined to performance fluctuations under single visual perturbations.}

\subsection{Limitations of Current Robotic Manipulation Benchmarks}

The above challenges indicate that robotic manipulation benchmarks for generalization and robustness diagnosis should satisfy \emph{\textbf{four requirements}}: \textcircled{1} fine-grained capability diagnosis that attributes model failures to specific stages along the perception, understanding, reasoning, and execution pipeline; \textcircled{2} unified evaluation of zero-shot generalization under both in-distribution (ID) and out-of-distribution (OOD) scenarios; \textcircled{3} systematic robustness testing under representative perturbations drawn from real-world deployment conditions; and \textcircled{4} a closed-loop mechanism that leverages diagnosis results to guide model refinement and re-evaluation.

\emph{\textbf{Long-horizon and knowledge-transfer robotic manipulation benchmarks}}. Benchmarks such as \emph{\textsc{Libero}}~\cite{liu2023libero}, \emph{CALVIN}~\cite{mees2022calvin}, \emph{RLBench}~\cite{james2020rlbench}, and \emph{VLABench}~\cite{zhang2025vlabench} provide unified environments for model comparison through standardized task suites, long-horizon manipulation procedures, or knowledge-transfer settings. \uline{However, they mainly focus on overall completion rates over predefined tasks, with limited fine-grained failure diagnosis and systematic ID/OOD scenario splits. As a result, it remains difficult to assess the true generalization ability of models under unseen tasks and complex scenario variations.}

\emph{\textbf{Generalization-driven robotic manipulation benchmarks}}. Benchmarks such as \emph{REALM}~\cite{sedlacek2026realm}, \emph{SimplerEnv}~\cite{li2024evaluating}, \emph{GemBench}~\cite{garcia2025towards}, \emph{AGNOSTOS}~\cite{zhou2026x-icm}, and \emph{Maniskill2}~\cite{gu2023maniskill2} extend the generalization evaluation from the perspectives of unseen objects, unseen environments, layout variations, and \emph{Sim2Real} transfer. 
\uline{However, these benchmarks primarily evaluate distribution shifts at the task or environment level, without a unified ID/OOD evaluation across visual perception, instruction understanding, spatial reasoning, and action execution. This makes it difficult to pinpoint the root causes of model degradation in OOD scenarios or to support subsequent refinement and re-evaluation.}

\emph{\textbf{Robustness and perturbation robotic manipulation benchmarks}}. Benchmarks such as \textsc{Libero-Plus}~\cite{fei2026libero-plus}, \emph{LIBERO-Pro}~\cite{zhou2025libero-pro}, \emph{VLA-Arena}~\cite{zhang2025vla-arena}, \emph{COLOSSEUM}~\cite{pumacay2024colosseum}, \emph{VLATest}~\cite{wang2025vlatest}, and \emph{LIBERO-X}~\cite{wang2026libero-x} extend robustness evaluation through perturbations in object layouts, visual appearances, camera viewpoints, and language expressions. VLATest focuses on automatic test generation, while \textsc{Libero-Plus} and COLOSSEUM attempt to connect perturbation evaluation with model improvement. \uline{However, most benchmarks remain limited to static or single perturbations, failing to adequately cover dynamic or combinatorial perturbations as encountered in real deployments. They offer no unified mechanism to translate perturbation results into actionable diagnostic optimization.}

Table~\ref{comparison} systematically compares these benchmarks. While they collectively cover long-horizon manipulation, knowledge transfer, in-distribution generalization, and perturbation robustness, each addresses only a subset of these dimensions. \uline{More critically, none provides fine-grained diagnosis of the unstable mapping among vision-language understanding, action representation, and control execution in VLA models, nor unified in-distribution versus out-of-distribution analysis across capability axes. Moreover, existing benchmarks offer little support for linking single-factor and combinatorial perturbation evaluation with capability-level diagnosis, targeted model optimization, and re-evaluation on held-out data.} \textbf{In contrast, \textsc{VLA-Gauge} is the first systematic framework that addresses all limitations together through a unified workflow of evaluation, diagnosis, model optimization, and re-evaluation. Spanning 6,300 task instances and 3,089 objects, the framework unifies three critical assessment axes: in-distribution versus out-of-distribution generalization, robustness under both isolated and combinatorial perturbations, and iterative re-evaluation driven by optimization feedback. This integrated design enables fine-grained, dimension-level diagnosis of model failures across various capability axes and perturbative conditions, while simultaneously providing actionable insights that directly inform and guide subsequent policy optimization.}
\section{Preliminaries}

\begin{figure*}[t]
\centering
\vspace{-3mm}
\includegraphics[width=0.9\textwidth]{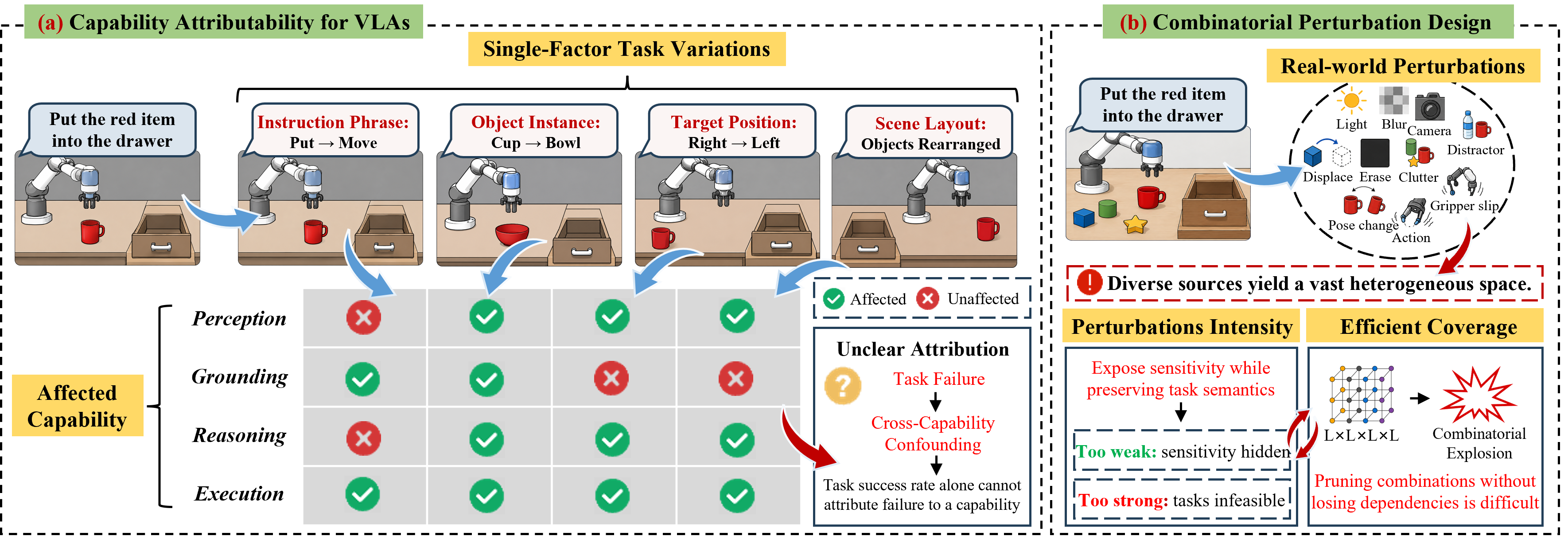}
\caption{\textbf{Examples Illustrating Challenges in VLA Generalization and Robustness Evaluation}}
\label{fig:challenge}
\vspace{-2.0em}
\end{figure*}

\subsection{\textbf{Capability Attributability for VLA Generalization Evaluation}}
\label{subsec:challenge1}

Given a task instruction and current observations, a VLA policy generates action commands through continuous interaction with the environment. Consequently, task completion relies on a multidimensional set of capabilities: visual perception, semantic grounding, world knowledge, spatial reasoning, action reasoning, and their synergistic coordination. Task-level success rates aggregate the requirements of these capabilities into a single metric, which fails to indicate which dimension of capability supports success or to pinpoint which dimension is responsible for failure. Moreover, imitation learning~\cite{hussein2017imitation} training may cause the policy to exploit recurring correlations between visual observations and action trajectories. Therefore, high performance on familiar tasks does not sufficiently demonstrate the model's capability generalization.

\uline{Multidimensional capability evaluation demands diagnostic attributability.} We consider a capability dimension to be attributable when performance variance across a task suite can be reliably traced to predefined variations in that specific dimension. To enable such interpretation, task objectives, success criteria, motor demands, and other extraneous factors must be held as consistent as possible across comparisons. Simply scaling up the number of tasks or altering surface-level visual attributes does not ensure attributability. Instead, the evaluation suite should be structured around explicit diagnostic targets, with clear specifications on which variables are manipulated and which are controlled.

\uline{However, establishing such attributability is inherently difficult in executable manipulation tasks, where success depends on a joint interplay of language instructions, object states, spatial relations, and action constraints.} Critically, as illustrated in Figure~\ref{fig:challenge}(a), varying a single factor, whether it be an instruction phrase, an object instance, a target position, or a layout configuration, can inadvertently shift multiple capability requirements simultaneously. This creates a fundamental tension: each capability dimension requires sufficient task variation to support generalization testing, yet introducing too much variation across tasks weakens within-suite comparability and invites cross-capability confounding. Conversely, keeping tasks too similar limits diagnostic coverage. Navigating this trade-off is a central benchmark-design challenge.

\vspace{-2mm}
\begin{tcolorbox}[
  colback=gray!10,
  colframe=black,
  boxrule=0.5pt,
  arc=2mm,
  left=1mm,
  right=1mm,
  top=0mm,
  bottom=0mm,
  enhanced
]
\small

\textcolor{red}{\faEdit}~\textbf{\emph{Challenge \#1}:} 
\emph{\textbf{Diagnostically Attributable Task Construction}.  A diagnostic benchmark must construct executable task suites that vary a predefined capability-related factor while preserving task goals, success conditions, and non-target requirements as consistently as possible. The central challenge is to maintain sufficient within-suite diversity without introducing cross-capability confounding, so that observed performance differences can be reliably attributed to a specific generalization bottleneck.}
\end{tcolorbox}
\vspace{-1.3em}
\subsection{\textbf{Combinatorial Perturbation Design for VLA Robustness Evaluation}}
\label{subsec:combinatorial_perturbations_design}

In real-world deployment, VLA policies operate within a closed loop of perception, decision, action, and state transition. A perturbation at any step, whether sensor noise, an unexpected object movement, or actuator deviation, can alter not only the immediate action but also subsequent environment states and future observations. For instance, an early perception bias may lead to misestimation of the target state, which in turn shifts the entire subsequent action trajectory. As a result, the impact of a perturbation can propagate and accumulate over the course of task execution. Open-loop evaluations on static inputs capture only the immediate, single-step reaction to a perturbation, missing these cascading effects. Robustness therefore must be assessed within the full interactive loop, where propagation and accumulation can be observed.

Single-factor tests remain useful for identifying local sensitivities and calibrating perturbation intensities. \uline{However, they do not characterize model behavior when multiple disturbances jointly affect perception, spatial configuration, and control execution.} We refer to these conditions as \emph{\textbf{combinatorial perturbations}}. Benchmarking combinatorial perturbations alongside single-factor baselines exposes robustness limits that isolated tests often miss.

We face a central challenge in constructing a diagnostic suite of combinatorial perturbations: four interdependent dimensions must be simultaneously balanced — realism, intensity, confound control, and combinatorial tractability: 
\begin{itemize}[leftmargin=*, itemsep=0.2em, topsep=0.2em]
    \item\textbf{\emph{Realism.}} As illustrated in Figure~\ref{fig:challenge}(b), perturbations in deployment originate from diverse sources, including sensor noise, environmental variations, and actuation errors, each with distinct parameter ranges. These perturbations must remain realistic while upholding the original task goal, motor requirements, and success criteria.
    \item\textbf{\emph{Appropriate intensity.}} Perturbation magnitudes should strike a balance between being large enough to elicit sensitivity and small enough to preserve task feasibility and avoid floor effects.
    \item\textbf{\emph{Controlled confounding.}} The task goal, motor requirements, and success conditions must remain consistent across perturbed and unperturbed settings to ensure that observed performance differences are attributable to the perturbations themselves.
    \item\textbf{\emph{Tractable combinatorial coverage.}} The space of perturbation combinations must be compressed while preserving cross-category dependencies; otherwise, exhaustive enumeration becomes computationally prohibitive, and the results resist meaningful interpretation.
\end{itemize}

\vspace{-1.0em}
\begin{tcolorbox}[
  colback=gray!10,
  colframe=black,
  boxrule=0.5pt,
  arc=2mm,
  left=1mm,
  right=1mm,
  top=0mm,
  bottom=0mm,
  enhanced
]
\small

\textcolor{red}{\faEdit}~\textbf{\emph{Challenge \#2}:}
\emph{\textbf{Diagnostic Combinatorial Perturbation Design}. A robustness benchmark must select perturbation factors, calibrate their intensities, and construct multi-factor configurations that preserve task feasibility and task semantics. The central challenge is to prune the combinatorial space without discarding plausible cross-category dependencies, so that the selected configurations remain computationally tractable, diagnostically interpretable, and representative of real-world deployment.}
\end{tcolorbox}
\section{VLA-Gauge Framework}

\begin{figure*}[t]
\centering
\vspace{-5mm}
\includegraphics[width=0.9\textwidth]{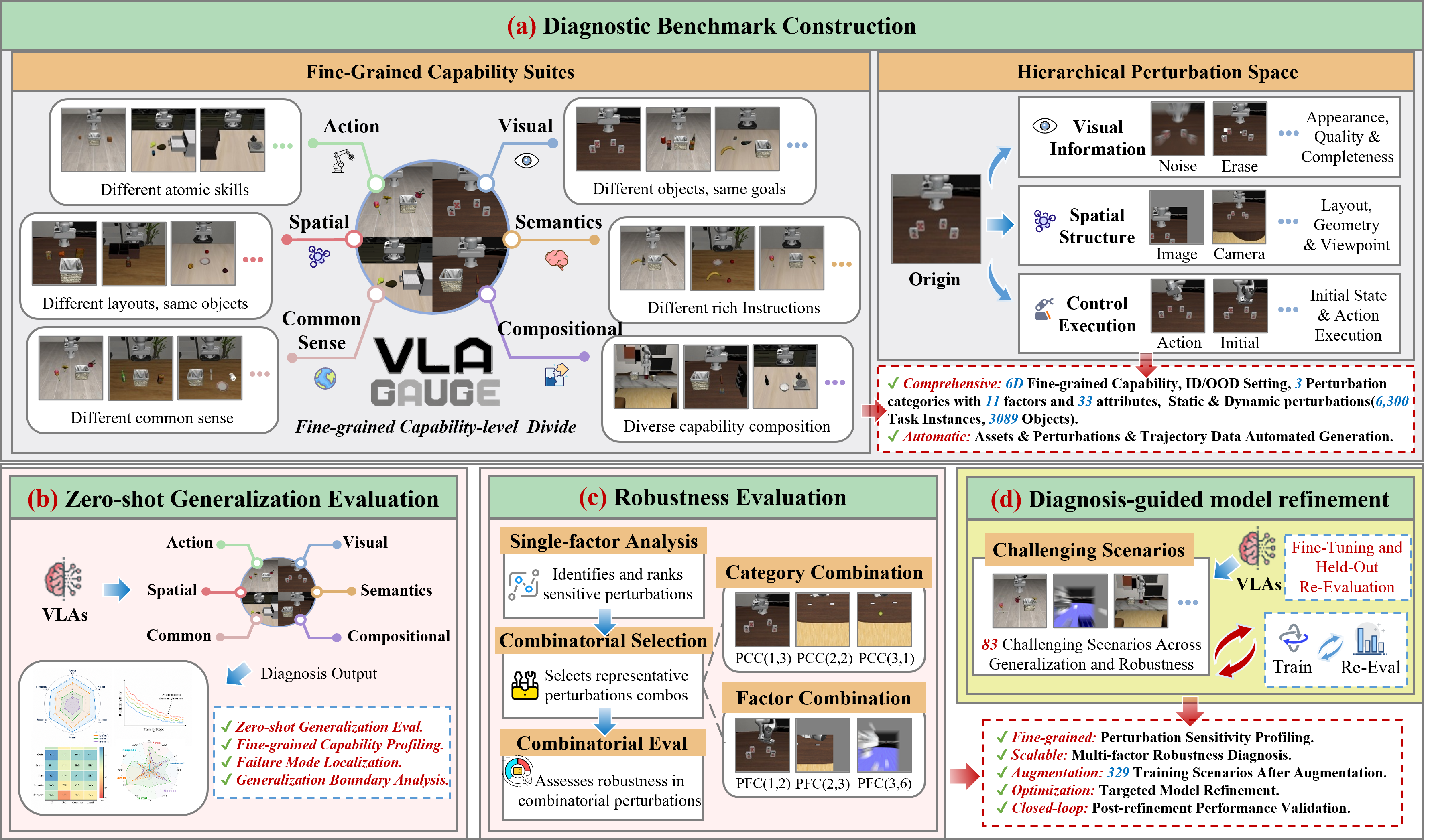}
\caption{\textbf{Overview of the \textsc{VLA-Gauge} Evaluation Framework}}
\label{fig:VLA-GAUGE}
\vspace{-1.2em}
\end{figure*}

We introduce \textsc{VLA-Gauge}, the first unified diagnostic framework for systematically evaluating zero-shot generalization and perturbation robustness in VLA models. Figure~\ref{fig:VLA-GAUGE} illustrates its architecture, which comprises four integrated stages: \textcircled{1} \emph{\textbf{benchmark construction}}, \textcircled{2} \emph{\textbf{zero-shot generalization evaluation}}, \textcircled{3} \emph{\textbf{robustness evaluation}}, and \textcircled{4} \emph{\textbf{diagnosis-driven model refinement}}.
The framework organizes base scenarios into capability-oriented suites with tailored ID and OOD settings. Evaluating models across these suites yields fine-grained capability profiles and ID-to-OOD generalization gaps, rather than a single aggregate success rate. These diagnostic insights then inform robustness testing under structured perturbations and guide targeted model refinement using verified successful demonstrations.
All components are built upon a unified executable task representation that encodes task semantics, execution conditions, and success criteria, enabling repeated closed-loop evaluations under capability-specific distribution shifts and diverse perturbation conditions.

Specifically, to address \emph{\textbf{Challenge \#1}}, \textsc{VLA-Gauge} organizes 270 base scenarios into six capability-oriented task suites, each with tailored ID and OOD conditions. Within each suite, variants are constructed around a primary diagnostic factor, while keeping the task goal, required motor skills, success criteria, and non-target conditions as consistent as possible. This design yields capability-level performance distributions and ID-to-OOD generalization gaps, rather than collapsing generalization into a single overall success rate. The resulting scenario-level performance also serves as a calibration anchor for subsequent robustness evaluation.

To address \emph{\textbf{Challenge \#2}}, \textsc{VLA-Gauge} compiles perturbation types and their value ranges that have been validated in existing studies~\cite{li2025cronusvla,zhang2025robustvla,zhang2025vla-arena,wang2026libero-x,zhou2025libero-pro,fei2026libero-plus,wang2025vlatest,
pumacay2024colosseum,xie2024decomposing,xing2021kitchenshift,chattopadhyay2021robustnav}, and organizes them according to their roles within the VLA execution pipeline. The resulting hierarchical framework encompasses three perturbation categories, namely visual information, spatial structure, and control execution, spanning 11 factors (e.g., camera, illumination, and background) and 33 perturbation attributes (e.g., orientation, distance, and position under camera perturbations). This framework captures the diverse and heterogeneous sources of perturbations present in real-world environments.
Single-factor sensitivity analysis calibrates perturbation magnitudes within the feasible range of each task, and evaluates model sensitivity while keeping the task goal, motor requirements, and success criteria fixed. Guided by the analysis results, the screening of perturbation categories and factors retains a representative set of multi-factor combinations rather than exhaustively testing all possible permutations. The resulting configurations are then evaluated across all base scenarios and the full model set under closed-loop execution, thereby achieving controlled and reproducible coverage of diverse combinatorial perturbations.

Finally, \textsc{VLA-Gauge} identifies challenging scenarios in which all 12 models achieve zero success under the initial diagnostic rollout configurations. For the zero-shot generalization and combinatorial perturbation robustness branches, we separately split the corresponding scenarios into training and test subsets that share no tasks or scenario configurations. Guided by the diagnostic results of each branch, we then apply data augmentation to the training subsets only. From the resulting training scenarios, we collect demonstrations verified as successful. The target model is then fine-tuned on these demonstrations and evaluated on the corresponding held-out test subset to determine whether the weaknesses revealed by diagnosis have been effectively mitigated. This process of evaluation, optimization, and re-evaluation forms a closed-loop diagnostic pipeline that integrates weakness identification, targeted fine-tuning, and independent validation.

\subsection{Diagnostic Benchmark Construction}
\label{subsec:diagnostic_benchmark_construction}

This diagnostic benchmark is built upon a unified executable task representation, enabling fine-grained capability attribution in VLA model evaluation. To faithfully replicate the diverse conditions encountered by robots in real-world environments, the benchmark encompasses three evaluation settings: clean-task, single-factor perturbation, and combinatorial perturbations. As illustrated in Figure~\ref{fig:VLA-GAUGE}(a), the construction proceeds along two parallel branches. The first branch, \textbf{fine-grained capability suites}, defines capability-specific in-distribution and out-of-distribution tasks while rigorously controlling non-target variables. The second branch, \textbf{hierarchical perturbation space}, generates controlled single-factor and combinatorial perturbations anchored to the same base tasks. Together, these two branches form a unified diagnostic space that redefines model assessment, shifting the focus from aggregate success-rate comparisons to capability attribution and perturbation robustness diagnostics.

\textbf{\emph{Unified Executable Task Representation.}} Before constructing the two parallel branches described above, we first build a unified representation of their common foundation, namely the robot manipulation task. Following the general definitions established in previous work~\cite{brohan2022rt,kim2024openvla}, we represent each base task as \emph{\textbf{a quadruple}} $\mathcal{T}=(\ell,\mathcal{O},\mathcal{C},\mathcal{E})$, corresponding to the \textcircled{1} \emph{\textbf{instruction}}, \textcircled{2} \emph{\textbf{scenario layout}},  \textcircled{3} \emph{\textbf{success criteria}}, and \textcircled{4} \emph{\textbf{execution-condition configuration}}, respectively, thereby enabling direct execution and repeated instantiation of every benchmark instance in a robot simulation environment:

\begin{itemize}[leftmargin=*, itemsep=0.2em, topsep=0.2em]
    \item\textbf{\emph{Instruction.}} A natural-language instruction describing the robot's task objective.
    
    \item\textbf{\emph{Scenario layout.}} It specifies the initial states and spatial relationships of manipulable, articulated, and fixed objects.
    
    \item\textbf{\emph{Success criteria.}} These define the logical conditions used to determine whether the task has been successfully completed.
    
    \item\textbf{\emph{Execution-condition configuration.}} It specifies environmental attributes (e.g., camera, light, and background) and robot attributes (e.g., gripper state and execution delay).
\end{itemize}

Although the Behavior Domain Definition Language (BDDL)~\cite{liu2023libero} can describe task objects, initial states, and goal conditions to construct base tasks, it has \emph{\textbf{two limitations}}: \textcircled{1} \emph{\textbf{Limited task representational scope}}, which does not fully encode execution conditions or environmental attributes; \textcircled{2} \emph{\textbf{Lack of unified task-variant modeling}}, which cannot generate controlled variants from base tasks while preserving task semantics and success conditions. Therefore, we introduce the Variant-augmented Behavior Domain Definition Language ($\mathrm{V\text{-}BDDL}$), which introduces an optional unified variant block (i.e., \texttt{:variant}). $\mathrm{V\text{-}BDDL}$ represents a task’s execution-condition configuration as \(\mathcal{E}=\{\rho_i\}_{i=1}^{n}\), with each variable encoded as a structured perturbation item \(\rho_i=(c_i,f_i,u_i,v_i,\mu_i)\), where
\begin{itemize}[leftmargin=*, itemsep=0.2em, topsep=0.2em]
    \item \emph{\textbf{Category}} ($c_i$) is determined by the primary stage of the VLA pipeline affected by the perturbation (e.g., perception, reasoning, or execution).
    \item \emph{\textbf{Factor}} ($f_i$) denotes a specific perturbation source within a category (e.g., camera, noise, or image). 
    \item \emph{\textbf{Attribute}} ($u_i$) denotes a configurable attribute of a perturbation factor (e.g., orientation, distance, or position under the camera factor).
    \item \emph{\textbf{Parameter value}} ($v_i$) specifies a perturbation attribute’s concrete setting (e.g., angular values from $2^\circ$ to $10^\circ$ for the yaw, pitch, and roll components of camera orientation).
    \item \emph{\textbf{Static and dynamic perturbations}} ($\mu_i$) specify how a perturbation item varies during task execution (e.g., static or dynamic).
\end{itemize} 

An empty execution-condition configuration \(\mathcal{E}=\varnothing\) denotes a clean base task, whereas configurations satisfying \(|\{f_i\}|=1\) or \(|\{f_i\}|>1\) instantiate single-factor or combinatorial perturbation variants, respectively. This design not only preserves BDDL’s original task representation capabilities but also unifies the modeling and instantiation of base and perturbed tasks, providing a unified executable task formalism for constructing fine-grained capability suites and a hierarchical perturbation space. Additionally, we convert assets from existing work~\cite{zhang2025vlabench,li2024evaluating,james2020rlbench,calli2015ycb,li2023behavior} into object definitions compatible with \textsc{Robosuite}~\cite{zhu2020robosuite}, forming a shared pool of 3,089 candidate objects. Building on these foundations, \textsc{VLA-Gauge} automates asset configuration and the generation of perturbations and trajectories, yielding 6,300 task instances across subsequent evaluations. Below we elaborate the construction of the two benchmark branches.

\mysubsubsection{Fine-Grained Capability Suite Construction.}
\label{subsec:tasks_sources_and_definition}

Considering the processing demands of robotic manipulation, including visual, language, and action modalities, as well as the semantic and world knowledge acquired during pretraining, we categorize VLA model capabilities into six dimensions: \textcircled{1} \emph{\textbf{visual perception}}, \textcircled{2} \emph{\textbf{semantic grounding}}, \textcircled{3} \emph{\textbf{common sense and world knowledge}}, \textcircled{4} \emph{\textbf{spatial reasoning}}, \textcircled{5} \emph{\textbf{action reasoning}}, and \textcircled{6} \emph{\textbf{compositional capability}}.
These dimensions systematically characterize the generalization of VLA models in closed-loop manipulation. Based on the unified executable task representation, we construct 270 clean base scenarios and organize them evenly into six fine-grained capability suites. This design uses the resulting performance profiles to attribute generalization failures to specific dimensions, thus revealing model weaknesses and informing subsequent diagnosis-driven refinement.

Each capability suite follows \emph{\textbf{a unified variable control principle}}: \textcircled{1} \emph{\textbf{Variation}}. Tasks are constructed around a single primary diagnostic variable, and only conditions directly related to this variable are modified. \textcircled{2} \emph{\textbf{Control}}. Non-target variables, including task objectives, success criteria, and other manipulation requirements, are kept as consistent as possible. This principle reduces confounding capabilities and attribution ambiguity, allowing performance differences to better reflect the target capability.

Following the variable control principle above, the six capability suites are constructed as follows:

\begin{itemize}[leftmargin=*, itemsep=0.2em, topsep=0.2em]
    \item \emph{\textbf{Visual perception}}. This suite varies object category, shape, texture, and visually similar distractors to assess whether models can identify, localize, and distinguish task-relevant objects from visual observations.
    \item \emph{\textbf{Semantic grounding}}. This suite varies linguistic cues (e.g., referring expressions, contextual descriptions, and implicit intents) to assess whether models can interpret diverse expressions and map them to executable goals and manipulation targets.
    \item \emph{\textbf{Common sense and world knowledge}}. This suite varies required knowledge types (e.g., object functions, category relations, everyday uses, and conventions) to assess whether models can infer unstated targets or intents using knowledge acquired during pretraining.
    \item \emph{\textbf{Spatial reasoning}}. This suite varies object positions, relative distances, directional relations, and reference object configurations to assess whether models can interpret scenario layouts and spatial relations to determine target objects and task constraints.
    \item \emph{\textbf{Action reasoning}}. This suite varies atomic skill requirements (e.g., pick\&place, open\&close, twist, and stack) and local action dependencies, while limiting subgoals and sequence length, to assess whether models can select and sequence appropriate skills.
    \item \emph{\textbf{Compositional capability}}. This suite varies subgoal count, skill combinations, and long-horizon dependencies across task stages to assess whether models can coordinate multiple skills across subgoals during multistep execution.
\end{itemize}

Compared with existing benchmarks with coupled capability requirements, this fine-grained decomposition \textcircled{1} more directly attributes generalization failures to individual capabilities and \textcircled{2} uses capability-specific clean-task baselines to distinguish underlying capability deficits from sensitivity to single and combinatorial perturbations.

Within each capability suite, we adopt the \textsc{Libero} training distribution shared by all evaluated models as the reference distribution, and build two evaluation settings on top of it.

\begin{itemize}[leftmargin=*, itemsep=0.2em, topsep=0.2em]
    \item \emph{\textbf{In-distribution (ID)}}. This setting preserves the task structures and object layout rules and adjusts the target diagnostic variable of the suite only within the value ranges and composition rules that this distribution already covers, so that it measures the zero-shot generalization of a model to unseen task instances drawn from the same distribution.
    \item \emph{\textbf{Out-of-distribution (OOD)}}. This setting, again guided by the diagnostic objective of each suite, introduces task structures, object categories, scenario layouts, instruction forms, or task compositions that the reference distribution does not cover, so that it measures the zero-shot generalization of a model to unseen task instances under these novel conditions.
\end{itemize}

The six suites thus yield 12 subsuites for comparing ID performance and ID-to-OOD generalization gaps across capabilities. Consequently, \textsc{VLA-Gauge} enables the identification of capability bottlenecks and the characterization of generalization decay across each capability from ID to OOD regimes, while maintaining a clean baseline without perturbations. This facilitates the downstream attribution of task failures to whether they stem from inherent capability limitations or from sensitivity to environmental perturbations.

\mysubsubsection{Hierarchical Perturbation Space Construction.}
\label{subsec:structured_perturbation_space}

In real-world deployment, we systematically screen perturbation factors that have been implemented and validated in existing studies of robot manipulation~\cite{li2025cronusvla},~\cite{zhang2025robustvla},~\cite{xie2024decomposing},~\cite{xing2021kitchenshift}, VLA evaluation~\cite{zhang2025vla-arena},~\cite{wang2026libero-x},~\cite{zhou2025libero-pro},~\cite{fei2026libero-plus},~\cite{wang2025vlatest},~\cite{pumacay2024colosseum} and embodied navigation~\cite{chattopadhyay2021robustnav}. Each factor must satisfy the following \emph{\textbf{selection criteria}}: \textcircled{1} it is reported in at least two studies; and \textcircled{2} it acts through a clearly identifiable visual, spatial, or control pathway. Based on the unified executable task representation, we encode each retained perturbation as a structured perturbation item $\rho_i \in \mathcal{E}$ and construct the hierarchical perturbation space of \textsc{VLA-Gauge}. This space contains \textbf{\emph{3 perturbation categories, 11 perturbation factors, 33 perturbation attributes, and 268 validated parameter values, while supporting static and dynamic perturbations as well as single and combinatorial perturbation configurations.}} Its hierarchical organization bounds the candidate space for subsequent combinatorial perturbation screening, reducing reliance on inefficient manual enumeration.

\begin{figure}[t]
\centering
\vspace{-0.6em}

\includegraphics[width=\linewidth]{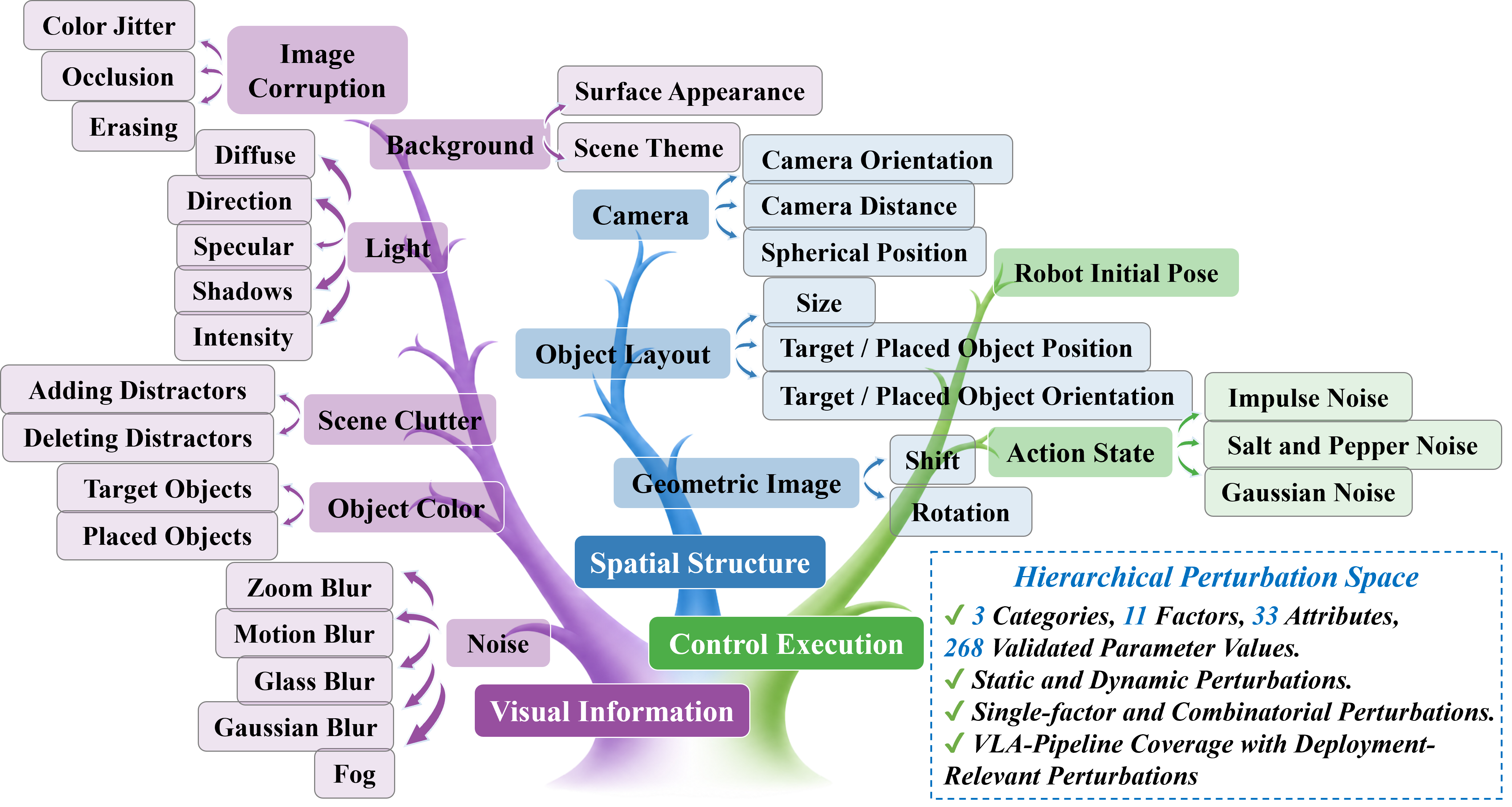}

\vspace{-0.6em}
\caption{Hierarchical Structure of the \textsc{VLA-Gauge} Perturbation Space}
\label{fig:perturbation_space}

\vspace{-1.4em}
\end{figure}

The perturbation space covers only execution-condition changes that preserve task semantics; linguistic variation and knowledge-based reasoning are assessed by the semantic grounding and common sense and world knowledge suites, respectively. Within this scope, we classify the perturbation items into three categories based on where they act in the closed loop of VLAs; their factors and perturbation attributes are shown in Figure~\ref{fig:perturbation_space}.


\begin{itemize}[leftmargin=*, itemsep=0.2em, topsep=0.2em]
    \item \emph{\textbf{Visual information perturbations}}. These perturbations target the appearance, quality, or integrity of visual observations to assess perceptual robustness to visual variation or degradation.
    \item \emph{\textbf{Spatial structure perturbations}}. These perturbations target object layouts, geometric relations, reference frames, or viewpoints to assess the robustness of a model’s spatial reasoning under changes in spatial configuration.
    \item \emph{\textbf{Control execution perturbations}}. These perturbations target initial robot states or action execution to assess a model’s closed-loop control robustness to state shifts or execution uncertainty.
\end{itemize}

Based on the parameter ranges reported in these studies, we generate candidate values for each perturbation attribute. Continuous parameter ranges (e.g., camera distance) are divided into subintervals and sampled uniformly within, whereas discrete or categorical parameter ranges (e.g., target objects or background) are sampled randomly from their value sets. Each candidate value is then instantiated in \textsc{Robosuite} and evaluated under \emph{\textbf{three validation criteria}}: \textcircled{1} \emph{\textbf{Perturbation effectiveness:}} the instantiated perturbation must alter the intended simulator, observation, or control property, with perceptible visual changes manually verified by comparing observations before and after injection. \textcircled{2} \emph{\textbf{Physical feasibility:}} the perturbed task must initialize correctly, maintain valid robot and object states and stable simulation, and remain executable. \textcircled{3} \emph{\textbf{Semantic consistency:}} the perturbation must preserve task instruction, goal predicates, required skill type, and success criteria. Only candidate values satisfying all three criteria are retained, thereby controlling confounding so that performance changes remain attributable to the perturbations themselves.

We further define static and dynamic perturbations \(\mu_i \in \{\mathrm{static}, \mathrm{dynamic}\}\) for each perturbation item \(\rho_i\): \textcircled{1} \emph{\textbf{Static perturbation}}. The value of the selected parameter is set at task initialization and remains fixed throughout execution. \textcircled{2} \emph{\textbf{Dynamic perturbation}}. The parameter value of the corresponding attribute is updated over time during execution.

To apply the set of execution conditions \(\mathcal{E}\) to a base task \(T\), we define a perturbation operator \(\mathcal{P}_{\mathcal{E}}\), which instantiates the perturbed task \(\widetilde{T}=\mathcal{P}_{\mathcal{E}}(T)\) via the $\mathrm{V\text{-}BDDL}$ (\texttt{:variant}) interface. The simulation parser implements this mapping by normalizing each perturbation item \(\rho_i \in \mathcal{E}\) and routing it to the corresponding simulation module according to its target. Thus, the hierarchical perturbation space converts heterogeneous real-world perturbations into reproducible perturbed tasks with controllable parameters. Compared with existing benchmarks that only evaluate perturbations in isolation, it supports systematic robustness diagnosis across perturbation categories and parameter settings, covering static and dynamic perturbations and their combinations.

\subsection{Benchmark Evaluation Protocol}
\label{subsec:diagnostic_evaluation_protocol}

To date, no existing VLA benchmark has combined the evaluation of zero-shot generalization and robustness under combinatorial perturbations within a single framework. \textsc{VLA-Gauge} addresses this gap through two complementary protocols: \textcircled{1} \emph{\textbf{Zero-shot generalization evaluation}}, assessing how well the model transfers its learned capabilities from ID to OOD settings and diagnosing generalization failures via paired ID/OOD tasks across six capability suites.
\textcircled{2} \emph{\textbf{Perturbation robustness evaluation}}, quantifying sensitivity to single-factor perturbations and robustness under combinatorial perturbations against clean-task baselines. Across both protocols, model parameters remain frozen during evaluation. Below, we elaborate on both protocol implementations.

\mysubsubsection{Zero-Shot Generalization Evaluation}
\label{subsec:generalization_evaluation}

As shown in Figure~\ref{fig:VLA-GAUGE}(b), \textsc{VLA-Gauge} evaluates the zero-shot generalization performance of each model across the six capability dimensions and reports its ID, OOD, and overall zero-shot score.

\begingroup
\setlength{\abovedisplayskip}{0pt}
\setlength{\belowdisplayskip}{0pt}
\setlength{\abovedisplayshortskip}{0pt}
\setlength{\belowdisplayshortskip}{0pt}

\vspace{-0.4em}
\begin{equation}
\mathrm{ZS}_{y}
=
\frac{1}{|\mathcal{D}|}
\sum_{d\in\mathcal{D}}
\mathrm{SR}_{y}(d),
\quad
y\in\{\mathrm{ID},\mathrm{OOD},\mathrm{ALL}\}
\label{eq:zero_shot_score}
\end{equation}

\endgroup

Here, \(\mathcal{D}\) denotes the set formed by the six capability dimensions, \(\mathrm{SR}_{y}\) denotes the model’s mean success rate under distribution \(y\), and \(\mathrm{ALL}=\mathrm{ID} \cup \mathrm{OOD}\). We evaluate each model with 50 closed-loop rollouts per scenario. All models share predefined initial states and random seeds and use the execution time limits specified by the corresponding suites. A rollout is successful if all goal predicates are satisfied within the time limit.

To reduce the cost of the subsequent perturbation robustness evaluation, we select the two models with the highest overall zero-shot score \(\mathrm{ZS}_{ALL}\) to form a fixed reference model set \(\mathcal{M}_{ref}\). Following \textsc{Libero-Plus}~\cite{fei2026libero-plus}, we classify a scenario as \textbf{\emph{Easy}}, \textbf{\emph{Medium}}, or \textbf{\emph{Hard}} according to whether both reference models, only one of them, or neither reaches the success rate threshold of 0.6. Within each difficulty level of every capability dimension, we take the scenarios whose average success rate over the two reference models lies closest to the median of that group as anchors, and these anchors constitute the anchor scenario set \(\mathcal S_{\mathrm{anchor}}\) used for calibration in the subsequent perturbation robustness evaluation.

\mysubsubsection{Perturbation Robustness Evaluation}
\label{subsec:perturbation_robustness_evaluation}

As shown in Figure~\ref{fig:VLA-GAUGE}(c), \textsc{VLA-Gauge} builds on the calibrated anchor scenarios and the hierarchical perturbation space, and carries out the evaluation in three successive stages: \textcircled{1} \emph{\textbf{Single-factor sensitivity analysis}}, which calibrates perturbation strength and identifies the perturbation factors to which a model is sensitive. \textcircled{2} \emph{\textbf{Combinatorial perturbation screening}}, which chooses representative perturbation combinations at both the category level and the factor level. \textcircled{3} \emph{\textbf{Combinatorial perturbations evaluation}}, which quantifies model robustness when several perturbation factors act jointly. All three stages adopt the $\mathrm{R\text{-}Score}$~\cite{li2025cronusvla} to measure how well a model preserves its performance under perturbation.

\refstepcounter{stage}
\noindent\textbf{\textit{Stage~\thestep: Single-Factor Sensitivity Analysis.}}
\label{stage:1}

During single-factor sensitivity analysis, \textsc{VLA-Gauge} evaluates the validated parameter values \(\mathcal V(u)\) for each attribute $u$ in turn on \(\mathcal{S}_{anchor}\) using \(\mathcal{M}_{ref}\), leaving all other attributes at their default values. It selects two types of sensitive values, which serve as the basis for sensitivity scoring and for ranking attributes and factors: \textcircled{1} \emph{\textbf{the overall degradation value set}} \(\mathcal{V}_{deg}\) contains candidate values that yield the lowest mean $\mathrm{R\text{-}Score}$ across the reference models, capturing the greatest overall performance degradation; and \textcircled{2} \emph{\textbf{the model-discriminative value set}} \(\mathcal{V}_{disc}\) contains candidate values that maximize the gap between the reference models’ mean $\mathrm{R\text{-}Scores}$ to distinguish their robustness responses.

\begingroup
\setlength{\abovedisplayskip}{0pt}
\setlength{\belowdisplayskip}{0pt}
\setlength{\abovedisplayshortskip}{0pt}
\setlength{\belowdisplayshortskip}{0pt}

\vspace{-0.4em}
\begin{equation}
\mathcal{V}_{\eta}(u)
=
\begin{cases}
\displaystyle
\underset{v\in\mathcal{V}(u)}{\arg\min}\,
\bar{R}(u,v),
& \eta=\mathrm{deg}, \\[6pt]
\displaystyle
\underset{v\in\mathcal{V}(u)}{\arg\max}\,
\Delta \bar{R}(u,v),
& \eta=\mathrm{disc}.
\end{cases}
\label{eq:factor_value_selection}
\end{equation}
\vspace{-0.2em}
\endgroup

Here, \(\bar R\) denotes the mean $\mathrm{R\text{-}Score}$ for the reference models on the fixed anchor set when attribute $u$ is set to $v$. \(\Delta\bar R\) denotes the absolute difference between the mean $\mathrm{R\text{-}Score}$ of the two reference models. All tied optimal values are retained, and the two sets of sensitive values are merged into \(\mathcal V_{\mathrm{sens}}\), which supports a unified sensitivity score and the subsequent \textbf{\emph{combinatorial perturbation screening}}. For each attribute, the $\mathrm{R\text{-}Scores}$ associated with its calibrated sensitive value set are averaged to obtain the attribute score, and the attribute scores within the same factor are then averaged to obtain the factor score. \textbf{\emph{Lower scores indicate greater perturbation sensitivity.}} These scores yield \textbf{\emph{three rankings}} for the subsequent screening phase: \textcircled{1} attributes within each factor, \textcircled{2} factors within each category, and \textcircled{3} factors globally across categories.

\refstepcounter{stage}
\noindent\textbf{\textit{Stage~\thestep: Combinatorial Perturbations Screening.}}
\label{stage:2}

Building on the calibrated sensitive value sets and the ranking produced in Stage~\ref{stage:1}, combinatorial perturbation screening proceeds at the category level and then at the factor level. Both levels adopt the same attribute configuration: each perturbation factor contributes the perturbation attribute with the lowest score, one value is drawn at random from the calibrated sensitive value set of that attribute and then held fixed throughout the screening process, and all remaining attributes stay at their default settings.

\textbf{\emph{Perturbation Category Screening.}} All perturbation category combinations (PCCs) formed from the three perturbation categories are enumerated. Within each combination, the lowest-scoring factor of every included category serves as the representative of that category, and the combination is instantiated with the attribute configuration described above. Candidate combinations that contain the same number of perturbation categories are then screened by the criterion given below, which yields the representative PCC set used for the subsequent factor-level screening.

\textbf{\emph{Perturbation Factor Screening.}} Within each retained PCC, the factor with the highest score, that is, the lowest sensitivity, is selected from every category to construct the initial perturbation factor combination (PFC). The remaining factors are then added one by one in descending order of factor score, which yields a nested sequence of candidates. The PFCs in each sequence are screened separately according to the criterion given below, and the retained combinations constitute the representative set of combinatorial perturbations in \textsc{VLA-Gauge}.

For a candidate perturbation combination \(\mathcal{E}\), the mean $\mathrm{R\text{-}Score}$ of reference model $m$ over the fixed scenario set \(\mathcal S_{\mathrm{anchor}}\) is converted into a degradation score \(D_\mathcal{E}(m)=\max(0,1-\bar{R}_\mathcal{E}(m)/100)\). A combination reaches the saturation boundary only when both reference models have a mean R-Score of zero, i.e., \(D_\mathcal{E}(m)=1\) for every \(m\in\mathcal M_{\mathrm{ref}}\). When a candidate combination is expanded, expansion stops as soon as this boundary is reached. Representative combinations are then selected using the criterion below. Following \emph{MMA}~\cite{yang2024mma}, we further define, for unsaturated candidates, \textbf{\emph{the Harmonic Mean of the Degradation Scores ($\boldsymbol{\mathit{D\text{-}HM}}$)}}:

\begingroup
\setlength{\abovedisplayskip}{0pt}
\setlength{\belowdisplayskip}{0pt}
\setlength{\abovedisplayshortskip}{0pt}
\setlength{\belowdisplayshortskip}{0pt}

\vspace{-0.4em}
\begin{equation}
\mathrm{D\text{-}HM}_\mathcal{E}
=
\frac{\left|\mathcal{M}_{ref}\right|}
{\displaystyle\sum_{m\in\mathcal{M}_{ref}}
\left(D_\mathcal{E}(m)+\epsilon\right)^{-1}}
-\epsilon
\label{eq:joint_degradation_score}
\end{equation}
\vspace{-0.2em}
\endgroup

Here, \(\epsilon=10^{-6}\). This metric is constrained by the smaller degradation score, so it favors combinations that cause substantial performance degradation for every reference model. Both screening levels retain the unsaturated candidate with the highest $\mathrm{D\text{-}HM}$, and if all candidates under consideration are saturated, the candidate with the highest mean degradation score is retained instead. All tied optimal candidates are kept.

\refstepcounter{stage}
\noindent\textbf{\textit{Stage~\thestep: Combinatorial Perturbations Evaluation.}}
\label{stage:3}

\textsc{VLA-Gauge} applies the representative PFCs selected in Stage~\ref{stage:2} to all base scenarios and reports the $\mathrm{R\text{-}Score}$ of each model under every PFC, which quantifies its robustness to combinatorial perturbations. Each model runs 50 closed-loop trials on each task instance, and every trial selects a static or a dynamic perturbation with equal probability.

Under static perturbation, every attribute retains throughout the rollout the pre-initialized value determined in Stage~\ref{stage:2}. Under dynamic perturbation, each mutable attribute that has at least two calibrated sensitive values resamples a value from its own calibrated sensitive value set at steps \(0\), \(\lfloor H_t/3\rfloor\), and \(\lfloor2H_t/3\rfloor\), where \(H_t\) denotes the maximum number of execution steps specified by the suite to which task \(t\) belongs. Attributes with only one calibrated sensitive value remain static throughout.

For a given PFC, base scenario, and rollout index, all models share the same perturbation mode (static or dynamic) and the same sequence of values. Sampling is independent across rollouts, and every perturbation configuration satisfies the semantic consistency requirements specified in~\ref{subsec:structured_perturbation_space}.

To quantify the overall performance degradation induced by each representative PFC, we build on \emph{RMSD}~\cite{sedlacek2026realm} and define the \textbf{\emph{Degradation Root Mean Square Deviation ($\boldsymbol{\mathit{D\text{-}RMSD}}$)}}:

\begingroup
\setlength{\abovedisplayskip}{0pt}
\setlength{\belowdisplayskip}{0pt}
\setlength{\abovedisplayshortskip}{0pt}
\setlength{\belowdisplayshortskip}{0pt}

\vspace{-0.4em}
\begin{equation}
\mathrm{D\text{-}RMSD}_{\mathcal{E}}
=
\sqrt{
\frac{1}{|\mathcal{M}|\,|\mathcal{S}_{\mathcal{E}|}}
\sum_{m\in\mathcal{M}}
\sum_{t\in\mathcal{S}_{\mathcal{E}}}
\left(
\Delta_\mathcal{E}(m, t)
\right)^2
}
\end{equation}
\vspace{-0.2em}
\endgroup

Here, \(\mathcal{M}\) denotes the full set of evaluated models, and \(\mathcal{S}_{\mathcal{E}}\) denotes the set of task instances generated by applying a representative PFC to the base scenarios. The quantity $\Delta_\mathcal{E}(m, t)
=\max(0,\mathrm{SR}_\varnothing(m, t)-\mathrm{SR}_\mathcal{E}(m, t))$ denotes the non-negative drop in success rate relative to the unperturbed condition. \textbf{\emph{A larger $\boldsymbol{\mathit{D\text{-}RMSD}}$ indicates a more severe overall performance degradation induced by that PFC.}}

\subsection{Diagnosis-Driven Model Refinement}
\label{subsec:diagnostic_guided_model_refinement}

\textbf{\emph{Data Construction.}} We use the capability diagnosis results from the zero-shot generalization branch and the combinatorial perturbation robustness branch to determine which tasks, scenario configurations, and perturbation states the optimization data needs to cover. For each of the two evaluation branches, we collect all scenarios in which every evaluated model achieves zero success under the initial diagnostic rollout configurations and organize them by the six capability dimensions, which yields two groups of challenging scenarios.  We then apply a fixed random seed to split each group into a training subset and a test subset that overlap in neither task nor scenario configuration, and we perform data augmentation on the training subset only.

In the zero-shot generalization branch, we retain the ID/OOD partition used in the evaluation and augment the challenging training scenarios to cover conditions associated with diagnosed capability weaknesses. Following the capability definitions in~\ref{subsec:tasks_sources_and_definition}, we construct variants using the following rules:

\begin{itemize}[leftmargin=*, itemsep=0.2em, topsep=0.2em]
    \item \emph{\textbf{Visual perception.}} We vary the target object's shape, color, or surface texture while keeping the manipulation goal unchanged. For instance, in a cup placement task, we change the cup's surface texture from a solid color to a striped pattern. If the instruction refers to the original texture, we update the corresponding description accordingly.
    \item \emph{\textbf{Spatial reasoning.}} We move either the target object or its movable reference object at a time, translating it to the left, to the right, toward the robot arm, or away from the robot arm while keeping all other objects in place. We generate four variants by moving the target object and four by moving the reference object, yielding eight variants per original scene. When necessary, we update the spatial descriptions in the instruction accordingly.
    \item \emph{\textbf{Semantic grounding.}} We rewrite instructions by adding contextual information, changing referring expressions, or conveying intent implicitly, while keeping the scene and task goal unchanged. For instance, we rewrite “Place the red cup in the tray” as “The red cup should be in the tray.”
    \item \emph{\textbf{Common sense and world knowledge.}} We construct object selection tasks that require identifying the target object based on its function, category membership, or physical properties. For instance, in a scene containing a spoon, a pen, and a toy, we use the instruction “Place the object used to eat soup in the tray.”
    \item \emph{\textbf{Action reasoning.}} We expand the range of manipulated object instances while keeping the atomic action type unchanged. For instance, in a drawer-opening task, we replace the drawer with another instance whose handle is in a different position.
    \item \emph{\textbf{Compositional capability.}} We modify the scene configurations and instructions of existing compositional tasks to create variants that require the joint use of multiple basic capabilities. For instance, in the task “Open the drawer and place the cup inside,” we position the target cup to the left of a bowl and revise the instruction to “Open the drawer and place the red cup located to the left of the bowl into the drawer.” The resulting task combines object recognition, spatial reasoning, and sequential manipulation.
\end{itemize}

In the combinatorial perturbation robustness branch, we select perturbation combinations based on the evaluation results. For each challenging training scenario, we use the parameter configuration retained during screening as the baseline. Each variant starts from this baseline configuration and changes only one attribute by randomly sampling a value different from the original from the corresponding subset of sensitive parameter values in~\ref{subsec:perturbation_robustness_evaluation}; all other attributes remain unchanged. For instance, given a combination of target object size, action noise, and image translation, we generate three new combinatorial perturbation configurations by changing each attribute once in a separate variant. We apply these configurations to the corresponding challenging training scenarios while keeping the task goals and success criteria unchanged.

Across the selected challenging training scenarios and their augmented counterparts, we collect the trajectories of the zero-shot generalization branch with a \emph{SpaceMouse}, and those of the combinatorial perturbation robustness branch through automated replay with the \textsc{VLA-Gauge} toolkit. The trajectories of a given task cover a variety of initial configurations and execution paths, which reduces the reliance of the model on a particular initial state or a fixed action pattern. After collection, task success is verified automatically against the goal predicate of each scenario, and trajectories that contain abnormal pauses, ineffective actions, or redundant paths are removed manually.

\textbf{\emph{Refinement Protocol.}} \textsc{VLA-Gauge} selects the model with the highest average performance on the zero-shot generalization and combinatorial perturbation robustness evaluations as the target model. The two optimization branches start from the same original checkpoint and train separate LoRA~\cite{hu2022lora} adapters on the generalization training set and the robustness training set respectively. The training setup and the LoRA hyperparameters follow the official configuration of the target model. A checkpoint is saved every 10,000 steps during training, and the test sets serve only for the final evaluation. For held-out evaluation, the original and fine-tuned models use a new, shared set of initial states and random seeds, independent of those used for scenario selection.



\section{Experiments}
We propose five research questions in our evaluation: 

\textcolor{green!75!black}{\faIcon{lightbulb}}~\emph{\textbf{RQ1 (Overall Zero-Shot Generalization Performance)}}: \emph{What is the overall zero-shot generalization performance of existing VLA models?}

\textcolor{green!75!black}{\faIcon{lightbulb}}~\emph{\textbf{RQ2 (Zero-Shot Performance Distribution and Capability Bottlenecks)}}: \emph{What zero-shot performance distribution and capability bottlenecks do existing VLA models exhibit across \textsc{VLA-Gauge}’s six capability dimensions?}

\textcolor{green!75!black}{\faIcon{lightbulb}}~\emph{\textbf{RQ3 (Sensitivity to Single-Factor Perturbations)}}: \emph{How sensitive are representative VLA models to different single-factor perturbations in \textsc{VLA-Gauge}?}

\textcolor{green!75!black}{\faIcon{lightbulb}}~\emph{\textbf{RQ4 (Robustness to Combinatorial Perturbations)}}: \emph{How robust are existing VLA models to combinatorial perturbations in \textsc{VLA-Gauge}?}

\textcolor{green!75!black}{\faIcon{lightbulb}}~\emph{\textbf{RQ5 (Effectiveness of Diagnosis-Driven Fine-Tuning)}}: \emph{To what extent does \textsc{VLA-Gauge}’s diagnosis-driven fine-tuning improve the target VLA model’s zero-shot generalization and combinatorial robustness?}

\subsection{Experiment Setup}



\emph{1) \textbf{Data Composition.}} \textsc{VLA-Gauge} primarily includes data for three stages: zero-shot generalization evaluation, perturbation robustness evaluation, and diagnosis-driven fine-tuning. The zero-shot generalization dataset comprises 270 unperturbed base scenarios across six capability suites. Each suite contains 45 scenarios categorized as ID or OOD. We select one easy, one medium, and one hard scenario from each suite, yielding 18 anchor scenarios.The robustness evaluation sequentially tests 268 validated parameter values across all attributes, followed by 7 PCCs and then 15 PFCs. The 3 selected representative PFCs are then applied to all base scenarios, yielding 810 task instances under combinatorial perturbations. During diagnosis-driven fine-tuning, we first select 27 challenging scenarios for zero-shot generalization and 56 for robustness to combinatorial perturbations. Following the train-test isolation setting in~\cite{toyer2020magical}, we partition the selected scenarios into training and test subsets at the scenario level, yielding 21 training and 6 test scenarios for the generalization branch and 45 training and 11 test scenarios for the robustness branch before augmentation. We collect 50 successful trajectories per training scenario, resulting in a total of 3,300 training demonstrations across the two branches before augmentation. We then apply diagnosis-driven data augmentation only to the training subsets. The augmentation methods and the final sizes of the training sets are detailed in~\ref{subsec:rq5}.

\emph{2) \textbf{Models and Metrics.}} We assess the zero-shot generalization and perturbation robustness of 12 VLA models using \textsc{VLA-Gauge}: \textcircled{1} \emph{\textbf{five large models}}, \emph{UniVLA}~\cite{bu2025univla}, \emph{OpenVLA-OFT}~\cite{kim2025fine}, \emph{RIPT-VLA}~\cite{tan2025interactive}, \emph{SimpleVLA-RL}~\cite{li2025simplevla} and \emph{CronusVLA}~\cite{li2025cronusvla}; \textcircled{2} \emph{\textbf{seven small models}}, \emph{$\pi_{0.5}$}~\cite{intelligence2025pi_}, \emph{Nora}~\cite{hung2025nora}, \emph{GR00T1.5}~\cite{nvidia2025gr00t}, \emph{DeepThinkVLA}~\cite{yin2025deepthinkvla}, \emph{X-VLA}~\cite{zheng2025x}, \emph{Evo-1}~\cite{lin2026evo} and \emph{VLA-Adapter-Pro}~\cite{wang2026vla}. These models span discrete, continuous, and hybrid action paradigms, as detailed in~\ref{subsec:baselines}.

We use five metrics: \textcircled{1} $\boldsymbol{\mathit{\%Success~Rate~(SR)}}$: the average binary success measure across all rollouts of each task; \textcircled{2} $\boldsymbol{\mathit{\%Zero\text{-}Shot~Score~(ZS)}}$: the mean performance across ID and OOD tasks within each suite, with both metrics assessing zero-shot generalization. \textcircled{3} $\boldsymbol{\mathit{R\text{-}Score~(R)}}$: performance retention relative to the unperturbed condition, used throughout robustness evaluation; \textcircled{4} $\boldsymbol{\mathit{D\text{-}HM}}$: cross-model consistency of performance degradation under combinatorial perturbations, used for perturbation screening. \textcircled{5} $\boldsymbol{\mathit{\%D\text{-}RMSD}}$: the overall degradation induced by combinatorial perturbations, used to select diagnosis-driven fine-tuning data. We evaluate fine-tuning with the corresponding evaluation-branch metrics.

\emph{3) \textbf{Evaluation Configuration.}} The evaluation follows the protocol in~\ref{subsec:diagnostic_evaluation_protocol}. For the initial zero-shot generalization and perturbation robustness evaluations, we use the officially released model checkpoints without additional training or adaptation. For each scenario, all models share predefined initial states and random seeds and perform 50 closed-loop rollouts. Each suite’s rollout horizon is set to 1.1 times the average human completion steps following \emph{LIBERO-X}~\cite{wang2026libero-x}.

Following the model-selection criterion in~\ref{subsec:diagnostic_guided_model_refinement}, we selected $\pi_{0.5}$ as the fine-tuning target and trained separate LoRA adapters for the generalization and robustness branches. Training was conducted on eight \texttt{NVIDIA A100 80GB GPUs}. The training configuration included: \textcircled{1} \emph{Action horizon}: 10; \textcircled{2} \emph{Batch size}: 4; \textcircled{3} \emph{Training duration}: 5 epochs; \textcircled{4} \emph{Learning rate}: 5e-5 initial rate with cosine decay; \textcircled{5} \emph{Warmup ratio}: 3\%; \textcircled{6} \emph{LoRA ranks}: 16 and 32 for the backbone and action expert, respectively.

\vspace{-1.2em}
\subsection{\textcolor{green!75!black}{\faIcon{lightbulb}}~RQ1: Overall Zero-Shot Generalization Performance}
\label{subsec:rq1}

\noindent\emph{\textbf{Methodology.}} We comprehensively assess the overall zero-shot generalization and cross-distribution performance gaps of VLA models. Following the unified evaluation protocol in~\ref{subsec:generalization_evaluation}, we keep model parameters frozen and evaluate 12 VLA models under in-distribution (ID) and out-of-distribution (OOD) conditions across the six capability suites of \textsc{VLA-Gauge}, and report the overall, ID, and OOD $\boldsymbol{\mathit{\%Zero\text{-}Shot~Score~(ZS)}}$ for each model. Under the ID condition, \textsc{VLA-Gauge} preserves the task structures, object layout rules, and instruction forms of the reference training distribution, and constructs unseen task instances within the existing value ranges and composition rules according to the evaluation requirements of the six capabilities. Under the OOD condition, \textsc{VLA-Gauge} introduces task structures, object categories, scenario layouts, instruction forms, or task compositions that the reference training distribution does not cover. The detailed settings of \textsc{VLA-Gauge} appear in ~\ref{subsec:tasks_sources_and_definition}. We also group models into large and small categories based on parameter count and compare their overall, ID, and OOD scores to examine the relationship between model size and zero-shot generalization performance on \textsc{VLA-Gauge}.

\begin{table}[t]
\centering
\caption{\textbf{Zero-shot generalization performance of VLA models on \textsc{VLA-Gauge}.}}
\label{tab:zero_shot_generalization}
\vspace{0.15em}

\begin{minipage}{0.98\columnwidth}
\centering
\scriptsize
\emph{
\textcolor{blue}{\textbf{Vis.}},
\textcolor{blue}{\textbf{Sem.}},
\textcolor{blue}{\textbf{CW.}},
\textcolor{blue}{\textbf{Spa.}},
\textcolor{blue}{\textbf{Act.}}, and
\textcolor{blue}{\textbf{Comp.}}
denote visual perception, semantic grounding, common sense and world knowledge, spatial reasoning, action reasoning, and compositional capability, respectively.
\textcolor{blue}{\textbf{ZS}} and
\textcolor{blue}{\textbf{Average}}
denote per-model means across capabilities and per-subset means across models, respectively.
All scores are percentages.
The \setlength{\fboxsep}{1pt}
\colorbox{bestbg}{\textcolor{blue}{top-1}} and
\colorbox{secondbg}{top-2}
results are highlighted, respectively.}
\end{minipage}

\vspace{0.25em}

\scriptsize
\setlength{\tabcolsep}{1.0pt}
\renewcommand{\arraystretch}{0.88}

\setlength{\aboverulesep}{0.22ex}
\setlength{\belowrulesep}{0.22ex}

\resizebox{\columnwidth}{!}{
\begin{tabular}{lcccccccc}
\toprule

\multicolumn{1}{c}{
    \multirow{2}{*}{\textbf{Models}}
}
&
\multirow{2}{*}{
    \emph{\textbf{Params}}
}
&
\multicolumn{6}{c}{
    \emph{\textbf{Generalization Capability}}
}
&
\multirow{2}{*}{
    \textcolor{blue}{\emph{\textbf{ZS}}}
}
\\

\cmidrule(lr){3-8}

& &
\textcolor{blue}{\emph{\textbf{Vis.}}}
&
\textcolor{blue}{\emph{\textbf{Sem.}}}
&
\textcolor{blue}{\emph{\textbf{CW.}}}
&
\textcolor{blue}{\emph{\textbf{Spa.}}}
&
\textcolor{blue}{\emph{\textbf{Act.}}}
&
\textcolor{blue}{\emph{\textbf{Comp.}}}
&
\\

\midrule
\rowcolor{gray!10}
\multicolumn{9}{c}{\textbf{Large Models}}
\\
\midrule

UniVLA~\cite{bu2025univla}
& 9B
& 29.3 & 18.7 & 23.6 & 10.2 & 35.9 & 23.6
& 23.6
\\

OpenVLA-OFT~\cite{kim2025fine}
& 7B
& 38.2 & 10.2 & 36.4 & 16.9 & 22.7 & 21.3
& 24.3
\\

RIPT-VLA~\cite{tan2025interactive}
& 7B
& 30.2 & 6.7 & 28.4 & 5.3 & 15.3 & 23.1
& 18.2
\\

SimpleVLA-RL~\cite{li2025simplevla}
& 7B
& 11.1 & 2.2 & 8.0 & 2.2 & 13.6 & 11.1
& 8.0
\\

CronusVLA~\cite{li2025cronusvla}
& 7B
& 24.4 & 28.9 & 29.3 & 8.4 & 18.2 & 22.7
& 22.0
\\

\midrule
\rowcolor{gray!10}
\multicolumn{9}{c}{\textbf{Small Models}}
\\
\midrule

$\pi$0.5~\cite{intelligence2025pi_}
& 3B
& \best{60.0}
& \best{57.8}
& \best{61.8}
& \best{47.6}
& \best{72.7}
& \best{41.8}
& \best{57.0}
\\

Nora~\cite{hung2025nora}
& 3B
& 11.1 & 9.3 & 8.4 & 2.2 & 6.8 & 1.8
& 6.6
\\

GR00T1.5~\cite{nvidia2025gr00t}
& 3B
& 15.1 & 16.0 & 17.3 & 4.9 & 14.5 & 17.3
& 14.2
\\

DeepThinkVLA~\cite{yin2025deepthinkvla}
& 2.9B
& \second{39.1}
& \second{39.6}
& \second{37.8}
& 21.8
& 33.2
& 29.8
& \second{33.6}
\\

X-VLA~\cite{zheng2025x}
& 0.9B
& 17.3
& 14.7
& 9.8
& \second{22.7}
& \second{37.3}
& \second{35.6}
& 22.9
\\

Evo-1~\cite{lin2026evo}
& 0.77B
& 30.2 & 28.0 & 30.7 & 8.0 & 28.2 & 23.6
& 24.8
\\

VLA-Adapter-Pro~\cite{wang2026vla}
& 0.5B
& 16.4 & 14.7 & 7.6 & 6.2 & 10.5 & 19.6
& 12.5
\\

\midrule
\textcolor{blue}{\emph{\textbf{Average}}}
& --
& \textbf{26.9}
& \textbf{20.6}
& \textbf{24.9}
& \textbf{13.0}
& \textbf{25.7}
& \textbf{22.6}
& \textbf{22.3}
\\

\bottomrule
\end{tabular}
}

\vspace{-2.7em}
\end{table}

\noindent\emph{\textbf{Results.}} Table~\ref{tab:zero_shot_generalization} shows that the 12 VLA models achieve overall zero-shot scores of \textcolor{blue}{6.6\%} to \textcolor{blue}{57.0\%} on \textsc{VLA-Gauge}, with a mean of only \textcolor{blue}{22.3\%}. \emph{$\pi_{0.5}$} ranks first at \textcolor{blue}{57.0\%}, while the remaining 11 models score no higher than \textcolor{blue}{33.6\%}. \emph{SimpleVLA-RL} achieves a zero-shot score of only \textcolor{blue}{8.0\%} on \textsc{VLA-Gauge}, ranking second to last among the evaluated models.

\textbf{\emph{Finding~1:}~The overall zero-shot generalization of the evaluated VLA models remains limited on the tasks covered by \textsc{VLA-Gauge}.} These results suggest that insufficient generalization may hinder reliable deployment in unseen environments.

Table~\ref{tab:id_ood_zero_shot} further compares the evaluated models' zero-shot performance under ID and OOD conditions in \textsc{VLA-Gauge}. Across the six capability suites, the 12 models achieve a mean ID score of only \textcolor{blue}{32.9\%} on unseen task instances. Under OOD conditions, the mean score drops further to \textcolor{blue}{7.2\%}, which is \textcolor{blue}{$\Downarrow$25.7} percentage points below ID score. In the 72 model-capability evaluations, \textcolor{blue}{71} yield lower OOD scores than the corresponding ID scores, and \textcolor{blue}{32} yield OOD scores of \textcolor{blue}{0}.

\textbf{\emph{Finding~2:}~Even when the task structures, layout rules, and instruction forms of the reference training distribution are preserved, the evaluated models still obtain low average ID zero-shot scores on the unseen task instances of \textsc{VLA-Gauge}, and their scores drop further under the OOD condition.} This highlights the need to assess generalization to unseen task instances within the reference distribution as well as under distribution shifts.

\begin{figure*}[t]
\centering

\begin{minipage}[t]{0.245\textwidth}
    \centering
    \includegraphics[width=\linewidth]{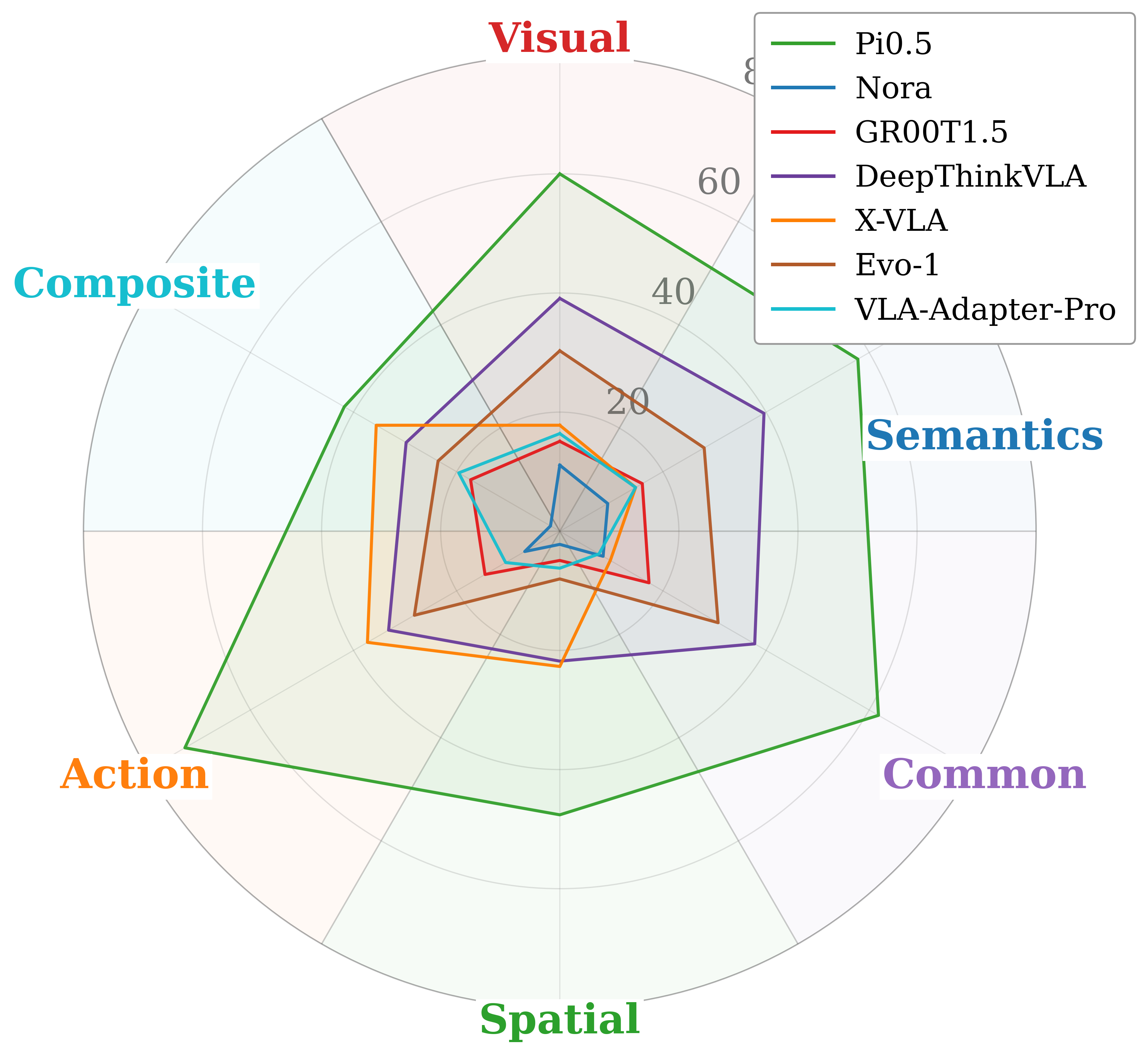}
    \vspace{-2mm}
    \makebox[\linewidth][c]{\small\textbf{(a)}}
\end{minipage}
\hfill
\begin{minipage}[t]{0.245\textwidth}
    \centering
    \includegraphics[width=\linewidth]{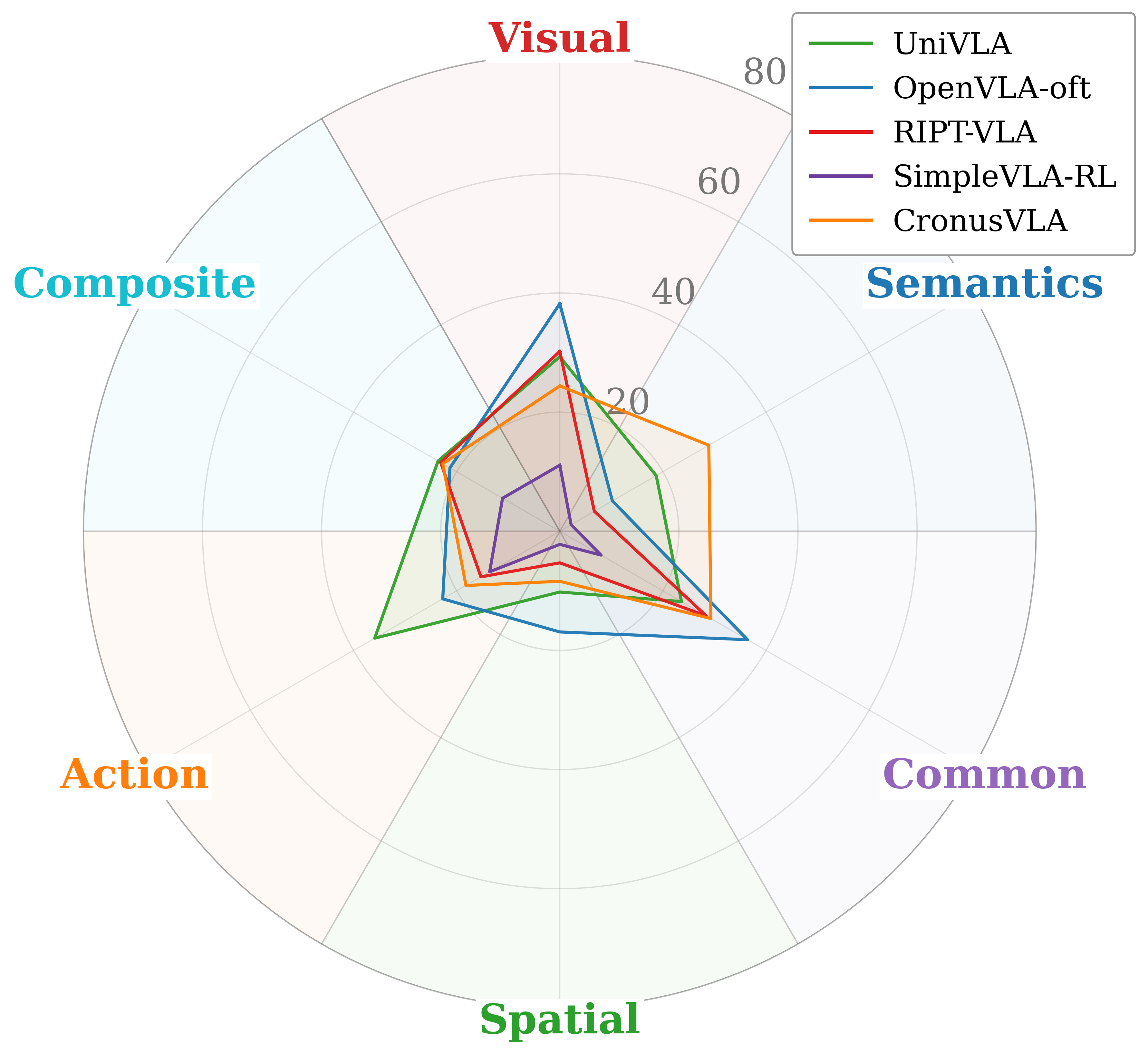}
    \vspace{-2mm}
    \makebox[\linewidth][c]{\small\textbf{(b)}}
\end{minipage}
\hfill
\begin{minipage}[t]{0.245\textwidth}
    \centering
    \includegraphics[width=\linewidth]{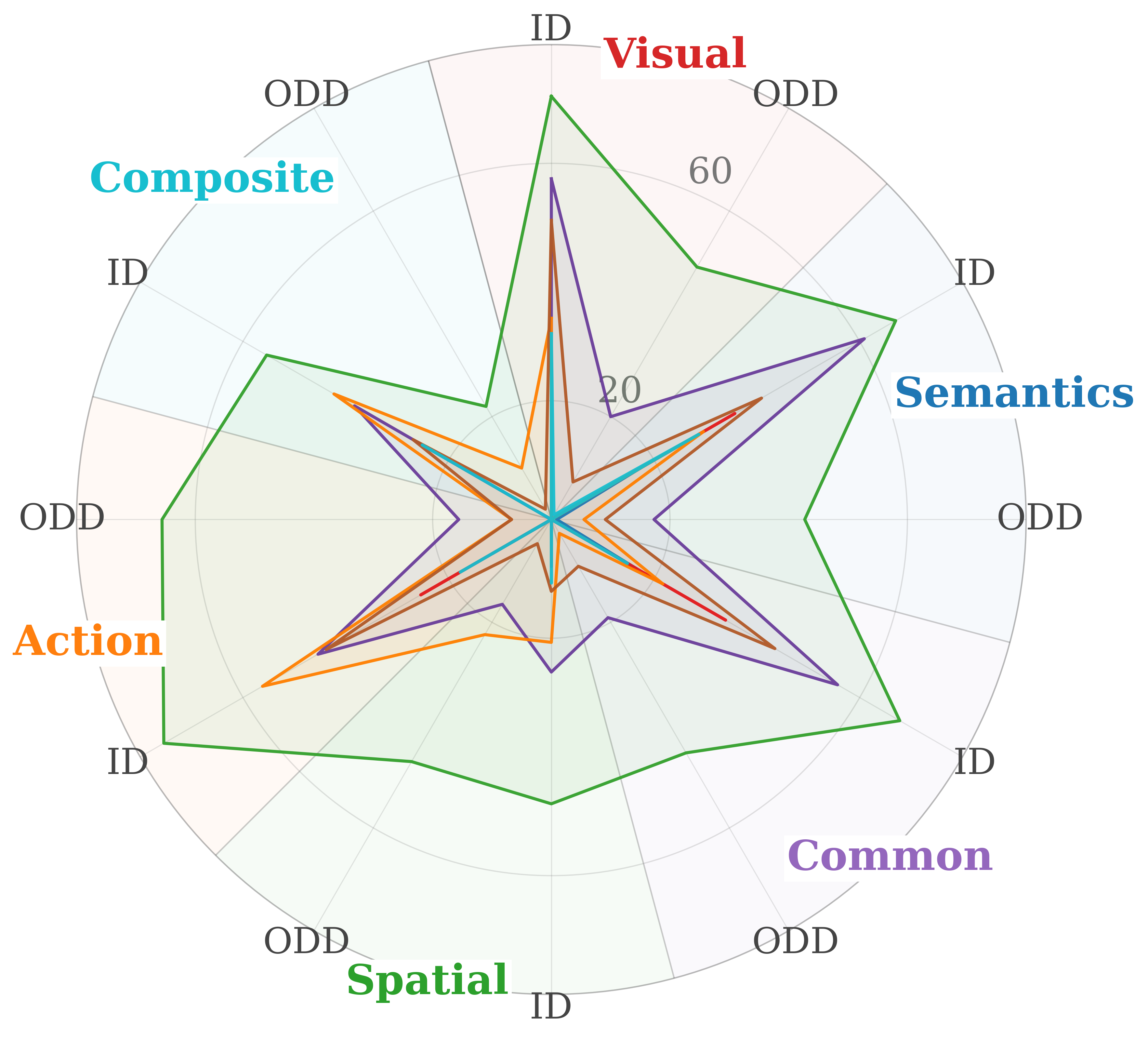}
    \vspace{-2mm}
    \makebox[\linewidth][c]{\small\textbf{(c)}}
\end{minipage}
\hfill
\begin{minipage}[t]{0.245\textwidth}
    \centering
    \includegraphics[width=\linewidth]{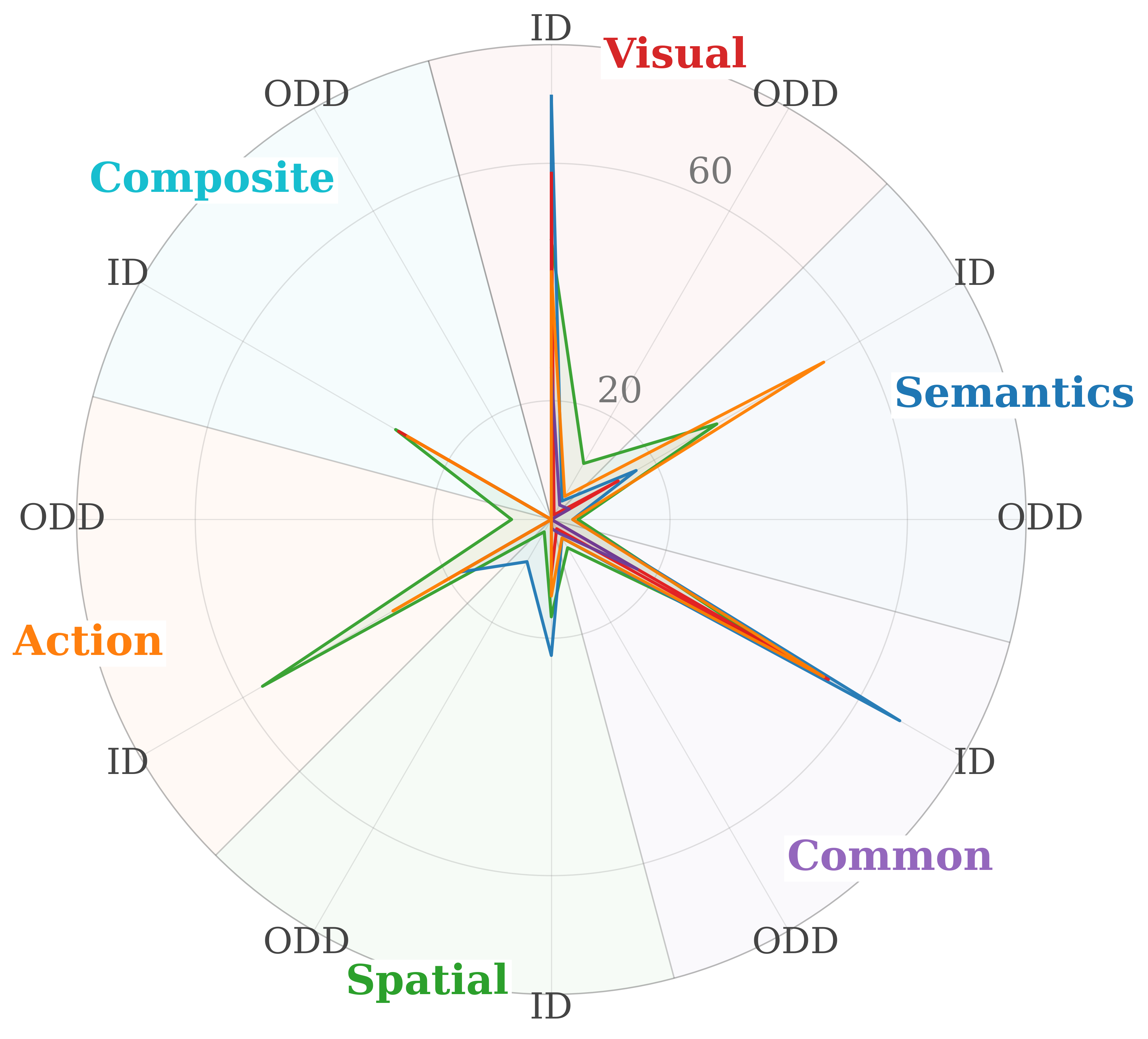}
    \vspace{-2mm}
    \makebox[\linewidth][c]{\small\textbf{(d)}}
\end{minipage}

\vspace{1mm}
\caption{\textbf{Radar distributions of VLA models across six capability dimensions.}
(a) Small-scale models.
(b) Large-scale models.
(c) ID/OOD performance of small-scale models.
(d) ID/OOD performance of large-scale models.}
\label{fig:radar_four}
\vspace{-1.5em}
\end{figure*}

As shown in Figure~\ref{fig:radar_four} and Table~\ref{tab:zero_shot_generalization}, the 3B-parameter \emph{$\pi$0.5} ranks first with an overall zero-shot score of \textcolor{blue}{57.0\%}, outperforming all five models with 7B to 9B parameters. Table~\ref{tab:id_ood_zero_shot} further shows that \emph{$\pi$0.5} achieves the highest mean scores under both ID and OOD conditions. One possible explanation is its joint training on heterogeneous data and its hybrid design that combines discrete representation learning with continuous action generation. Therefore, to improve zero-shot generalization, future work should consider broader task and scenario coverage in training data and more effective mechanisms for integrating visual, language, and action information, rather than relying solely on parameter scaling.

\textbf{\emph{Finding~3:}~Among the 12 models evaluated on \textsc{VLA-Gauge}, the zero-shot score does not improve consistently with parameter scale, so parameter scale alone is insufficient to explain the differences in zero-shot generalization across models.}

\begin{table*}[t]
\centering
\caption{\textbf{In-distribution and out-of-distribution zero-shot generalization performance across VLA models.}}
\label{tab:id_ood_zero_shot}
\vspace{0.15em}

\begin{minipage}{0.96\textwidth}
\centering
\small
\emph{
\textcolor{blue}{\textbf{ID/OOD}}:
in-distribution/out-of-distribution evaluation settings; The \setlength{\fboxsep}{1pt} \textcolor{blue}{\textbf{Average1.}} and \textcolor{blue}{\textbf{Average2.}} denote per-model means across capabilities and per-subset means across models. The \colorbox{bestbg}{\textcolor{blue}{top-1}} and \colorbox{secondbg}{top-2} results are highlighted, respectively. Near-zero ID/OOD scores appear in \textcolor{red}{red}. The overall results and the results specific to the ID/OOD settings are obtained from separate evaluation runs.}
\end{minipage}

\vspace{0.25em}

\scriptsize
\setlength{\tabcolsep}{4pt}        
\renewcommand{\arraystretch}{0.8}   

\setlength{\aboverulesep}{0.2ex}
\setlength{\belowrulesep}{0.2ex}

\resizebox{0.86\textwidth}{!}{%
\begin{tabular}{@{}l*{14}{c}@{}}
\toprule

\multicolumn{1}{c}{
    \multirow{2}{*}{\textbf{Models}}
}
& \multicolumn{2}{c}{
    \emph{\textbf{Vis.}}
}
& \multicolumn{2}{c}{
    \emph{\textbf{Sem.}}
}
& \multicolumn{2}{c}{
    \emph{\textbf{CW.}}
}
& \multicolumn{2}{c}{
    \emph{\textbf{Spa.}}
}
& \multicolumn{2}{c}{
    \emph{\textbf{Act.}}
}
& \multicolumn{2}{c}{
    \emph{\textbf{Comp.}}
}
& \multicolumn{2}{c}{
    \textcolor{blue}{\emph{\textbf{Average1.}}}
}
\\

\cmidrule(lr){2-3}
\cmidrule(lr){4-5}
\cmidrule(lr){6-7}
\cmidrule(lr){8-9}
\cmidrule(lr){10-11}
\cmidrule(lr){12-13}
\cmidrule(lr){14-15}

&
\textcolor{blue}{\emph{\textbf{ID}}}
& \textcolor{blue}{\emph{\textbf{OOD}}}
& \textcolor{blue}{\emph{\textbf{ID}}}
& \textcolor{blue}{\emph{\textbf{OOD}}}
& \textcolor{blue}{\emph{\textbf{ID}}}
& \textcolor{blue}{\emph{\textbf{OOD}}}
& \textcolor{blue}{\emph{\textbf{ID}}}
& \textcolor{blue}{\emph{\textbf{OOD}}}
& \textcolor{blue}{\emph{\textbf{ID}}}
& \textcolor{blue}{\emph{\textbf{OOD}}}
& \textcolor{blue}{\emph{\textbf{ID}}}
& \textcolor{blue}{\emph{\textbf{OOD}}}
& \textcolor{blue}{\emph{\textbf{ID}}}
& \textcolor{blue}{\emph{\textbf{OOD}}}
\\

\midrule
\rowcolor{gray!10}
\multicolumn{15}{c}{\textbf{Large Models}}
\\
\midrule

UniVLA~\cite{bu2025univla}
& 47.0 & 10.9
& 32.2 & 4.5
& 40.9 & 5.5
& 16.4 & 2.4
& \second{56.2} & 6.7
& 30.3 & \textcolor{red}{0.0}
& 37.2 & 5.0
\\

OpenVLA-OFT~\cite{kim2025fine}
& \best{71.3} & 3.6
& 16.5 & 3.6
& \best{67.8} & 3.6
& 22.9 & 8.2
& 17.7 & \textcolor{red}{0.0}
& 27.4 & \textcolor{red}{0.0}
& 37.3 & 3.2
\\

RIPT-VLA~\cite{tan2025interactive}
& \second{58.3} & \textcolor{red}{0.9}
& 13.0 & \textcolor{red}{0.0}
& 53.9 & 1.8
& 9.3 & \textcolor{red}{0.0}
& 20.8 & \textcolor{red}{0.0}
& 29.7 & \textcolor{red}{0.0}
& 30.8 & \textcolor{red}{0.5}
\\

SimpleVLA-RL~\cite{li2025simplevla}
& 25.4 & 2.8
& 3.5 & \textcolor{red}{0.0}
& 16.7 & 2.7
& 1.5 & \textcolor{red}{0.0}
& 15.4 & \textcolor{red}{0.0}
& 11.5 & \textcolor{red}{0.0}
& 12.3 & \textcolor{red}{0.9}
\\

CronusVLA~\cite{li2025cronusvla}
& 41.7 & 4.5
& 53.0 & 3.6
& 53.0 & 3.6
& 12.9 & \textcolor{red}{0.0}
& 30.8 & \textcolor{red}{0.0}
& 28.0 & \textcolor{red}{0.0}
& 36.6 & 2.0
\\

\midrule
\rowcolor{gray!10}
\multicolumn{15}{c}{\textbf{Small Models}}
\\
\midrule

$\pi$0.5~\cite{intelligence2025pi_}
& \best{71.3} & \best{49.1}
& \best{67.0} & \best{42.7}
& \best{67.8} & \best{46.4}
& \best{47.9} & \best{47.1}
& \best{75.4} & \best{65.6}
& \best{55.4} & \best{22.0}
& \best{64.1} & \best{45.5}
\\

Nora~\cite{hung2025nora}
& 21.7 & \textcolor{red}{0.0}
& 17.4 & \textcolor{red}{0.1}
& 16.5 & \textcolor{red}{0.0}
& 3.6 & \textcolor{red}{0.0}
& 11.5 & \textcolor{red}{0.0}
& 2.3 & \textcolor{red}{0.0}
& 12.2 & \textcolor{red}{0.0}
\\

GR00T1.5~\cite{nvidia2025gr00t}
& 30.4 & \textcolor{red}{0.0}
& 35.7 & \textcolor{red}{0.0}
& 33.9 & \textcolor{red}{0.0}
& 7.9 & \textcolor{red}{0.0}
& 25.4 & \textcolor{red}{0.0}
& 20.0 & \textcolor{red}{0.0}
& 25.6 & \textcolor{red}{0.0}
\\

DeepThinkVLA~\cite{yin2025deepthinkvla}
& 57.4 & \second{20.0}
& \second{60.9} & \second{17.3}
& \second{55.7} & \second{19.1}
& \second{25.7} & 16.5
& 45.4 & \second{15.6}
& 38.3 & \textcolor{red}{0.0}
& \second{47.2} & \second{14.8}
\\

X-VLA~\cite{zheng2025x}
& 33.9 & \textcolor{red}{0.0}
& 29.6 & 5.5
& 21.7 & 2.7
& 20.7 & \second{22.4}
& \second{56.2} & 6.7
& \second{42.3} & \second{10.0}
& 34.1 & 7.9
\\

Evo-1~\cite{lin2026evo}
& 50.4 & 7.3
& 40.9 & 9.1
& 43.5 & 9.1
& 12.1 & 4.7
& 43.8 & 6.7
& 26.9 & 2.0
& 36.3 & 6.5
\\

VLA-Adapter-Pro~\cite{wang2026vla}
& 31.3 & \textcolor{red}{0.9}
& 28.7 & \textcolor{red}{0.0}
& 14.8 & \textcolor{red}{0.0}
& 10.7 & \textcolor{red}{0.0}
& 17.7 & \textcolor{red}{0.0}
& 25.1 & \textcolor{red}{0.0}
& 21.4 & \textcolor{red}{0.2}
\\

\midrule
\textcolor{blue}{\emph{\textbf{Average2.}}}
& 45.0 & 8.3
& 33.2 & 7.2
& 40.5 & 7.9
& 16.0 & 8.4
& 34.7 & 8.4
& 28.1 & 2.8
& 32.9 & 7.2
\\

\bottomrule
\end{tabular}
}

\vspace{-1.5em}
\end{table*}

\vspace{-1.2em}
\subsection{\textcolor{green!75!black}{\faIcon{lightbulb}}~RQ2: Zero-Shot Performance Distribution and Capability Bottlenecks}
\label{subsec:rq2}

\noindent\emph{\textbf{Methodology.}}~To examine how generalization performance varies across the six capability dimensions of \textsc{VLA-Gauge} and identify capability bottlenecks, we use the unified evaluation results from~\ref{subsec:rq1} and the six capability suites defined in~\ref{subsec:tasks_sources_and_definition}. For each dimension, we compute the mean overall, ID, and OOD zero-shot scores across the 12 models and count the models with an OOD score of zero. We define the dimension with the lowest mean overall score as the overall capability bottleneck, and the dimension with both the lowest mean OOD score and the most models scoring zero under OOD conditions as the OOD capability bottleneck. Both bottlenecks represent performance weaknesses relative to the other capability dimensions within this evaluation. We measure performance degradation under OOD conditions as the mean ID score minus the mean OOD score for each dimension. As described in ~\ref{subsec:tasks_sources_and_definition}, the three suites for visual perception, semantic grounding, and common sense and world knowledge share scenario layouts and differ only in task instructions. We compare their mean overall, ID, and OOD scores to examine performance differences under different instruction requirements.

\noindent\emph{\textbf{Results.}} Table~\ref{tab:zero_shot_generalization} and Figure~\ref{fig:radar_four}(a)--(b) show clear differences in mean zero-shot performance across the six capability dimensions for the 12 models. Spatial reasoning has a mean zero-shot score of only \textcolor{blue}{13.0\%}, the lowest among the six dimensions and at least \textcolor{blue}{7.6} percentage points below that of any other dimension. For 8 of the 12 models, spatial reasoning is the lowest-scoring dimension or ties for the lowest score.

As shown in Table~\ref{tab:id_ood_zero_shot} and Figure~\ref{fig:radar_four}(c)--(d), spatial reasoning also has the lowest mean ID score among the six dimensions, only \textcolor{blue}{16.0\%}. This weakness may be related to most evaluated models' primary reliance on 2D RGB inputs and limited coverage of spatial relations and layout variations in the training data. Future work may prioritize improving spatial reasoning under ID conditions by drawing on existing methods for integrating depth maps and point clouds and expanding training data coverage of target object positions, directional descriptions, and variations in reference objects. These changes can be evaluated to test whether 3D information and spatial training data improve zero-shot spatial reasoning.

\textbf{\emph{Finding~4:}~Spatial reasoning constitutes the overall capability bottleneck of the evaluated VLA models, and this weakness is already evident under the ID condition.} With all other task conditions held fixed, the models perform poorly on spatial tasks that involve changes in the position of the target object, in the directional description within the instruction, and in the reference object.

Under OOD conditions, compositional capability has the lowest mean score, only \textcolor{blue}{2.8\%}. \textcolor{blue}{9} of the 12 models score zero in this dimension, the largest number across the six dimensions. In comparison, visual perception, semantic grounding, common sense and world knowledge, and action reasoning have higher mean OOD scores, but all remain in the single digits and are well below their corresponding ID scores. We further observe that common sense and world knowledge achieves a mean score close to that of visual perception under both ID and OOD conditions, whereas semantic grounding scores lower than both.

These patterns may reflect insufficient compositional task examples in the evaluated models' training data, along with limited coverage of long or implicitly phrased instructions and subtask execution. Future work may prioritize adding compositional task examples that cover different capability combinations and strengthening training on subtask execution. It can also strengthen training in understanding long or implicitly phrased instructions and add examples that require common sense and world knowledge to interpret task requirements and select appropriate actions, with the aim of improving zero-shot performance under OOD conditions. In the diagnosis-driven fine-tuning experiments in~\ref{subsec:rq5}, we further test whether this data augmentation approach improves zero-shot generalization. 

\textbf{\emph{Finding~5:}~Compositional capability constitutes the out-of-distribution capability bottleneck of the evaluated VLA models, whose OOD zero-shot performance is particularly weak on multi-step tasks that require coordinating several subgoals, combining different manipulation skills, and handling long-horizon dependencies across task stages.} Under the OOD condition, the models also show limited understanding of instructions that contain long descriptions or implicit expressions, together with limited ability to execute the corresponding subtasks.

\vspace{-1.2em}
\subsection{\textcolor{green!75!black}{\faIcon{lightbulb}}~RQ3: Sensitivity to Single-Factor Perturbations}
\label{subsec:rq3}

\noindent\emph{\textbf{Methodology.}}~Following the selection rules in~\ref{subsec:generalization_evaluation}, we use the zero-shot generalization results from~\ref{subsec:rq1} and ~\ref{subsec:rq2} to select the reference model set $M_{ref}=\{\pi_{0.5},\mathrm{DeepThinkVLA}\}$ and an 18-scenario anchor scenario set $S_{anchor}$, respectively. Under the single-factor evaluation protocol in~\ref{subsec:perturbation_robustness_evaluation}, we examine 33 perturbation attributes associated with 11 perturbation factors across three categories, covering 268 validated non-default parameter settings. We apply each setting to all 18 anchor scenarios, changing only the corresponding perturbation attribute while keeping all other attributes at their default values. This yields 4,824 perturbed tasks, each evaluated with both reference models.

We used each model's unperturbed performance in the same scenario as the baseline and quantified performance retention under perturbation using $\boldsymbol{\mathit{R\text{-}Score~(R)}}$. Lower $\mathrm{R\text{-}Score}$ values indicate greater sensitivity to the perturbation. Using this metric, we compare the magnitude of performance changes across perturbation attributes for the two reference models and examine how their performance varies with parameter settings.

\begin{figure*}[!t]
\centering

\begin{minipage}[t]{0.33\textwidth}
    \centering
    \includegraphics[width=\linewidth]{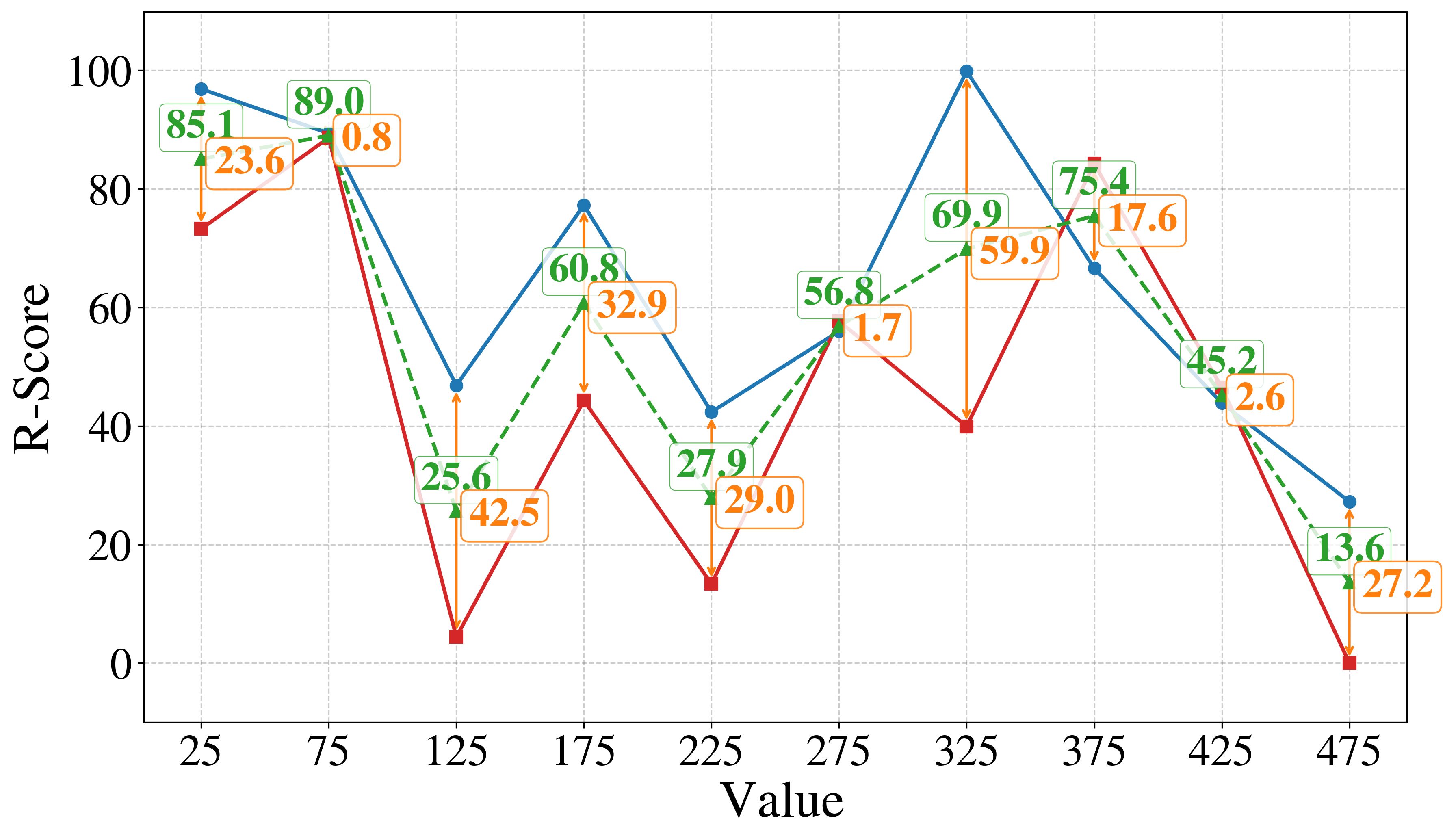}
    \par\vspace{-1mm}
    {\small\textbf{(a)}}\par
\end{minipage}%
\hfill
\begin{minipage}[t]{0.33\textwidth}
    \centering
    \includegraphics[width=\linewidth]{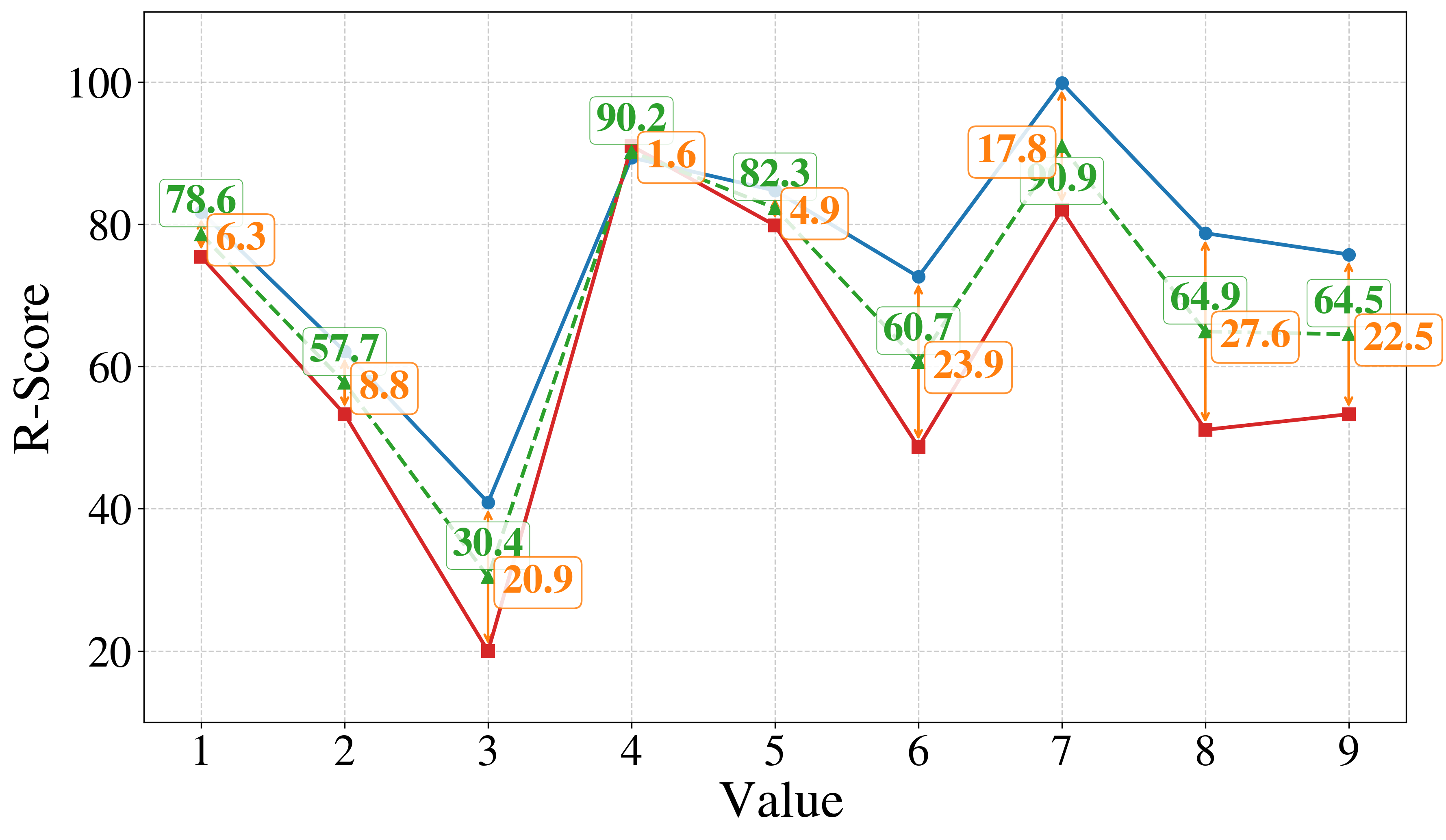}
    \par\vspace{-1mm}
    {\small\textbf{(b)}}\par
\end{minipage}%
\hfill
\begin{minipage}[t]{0.33\textwidth}
    \centering
    \includegraphics[width=\linewidth]{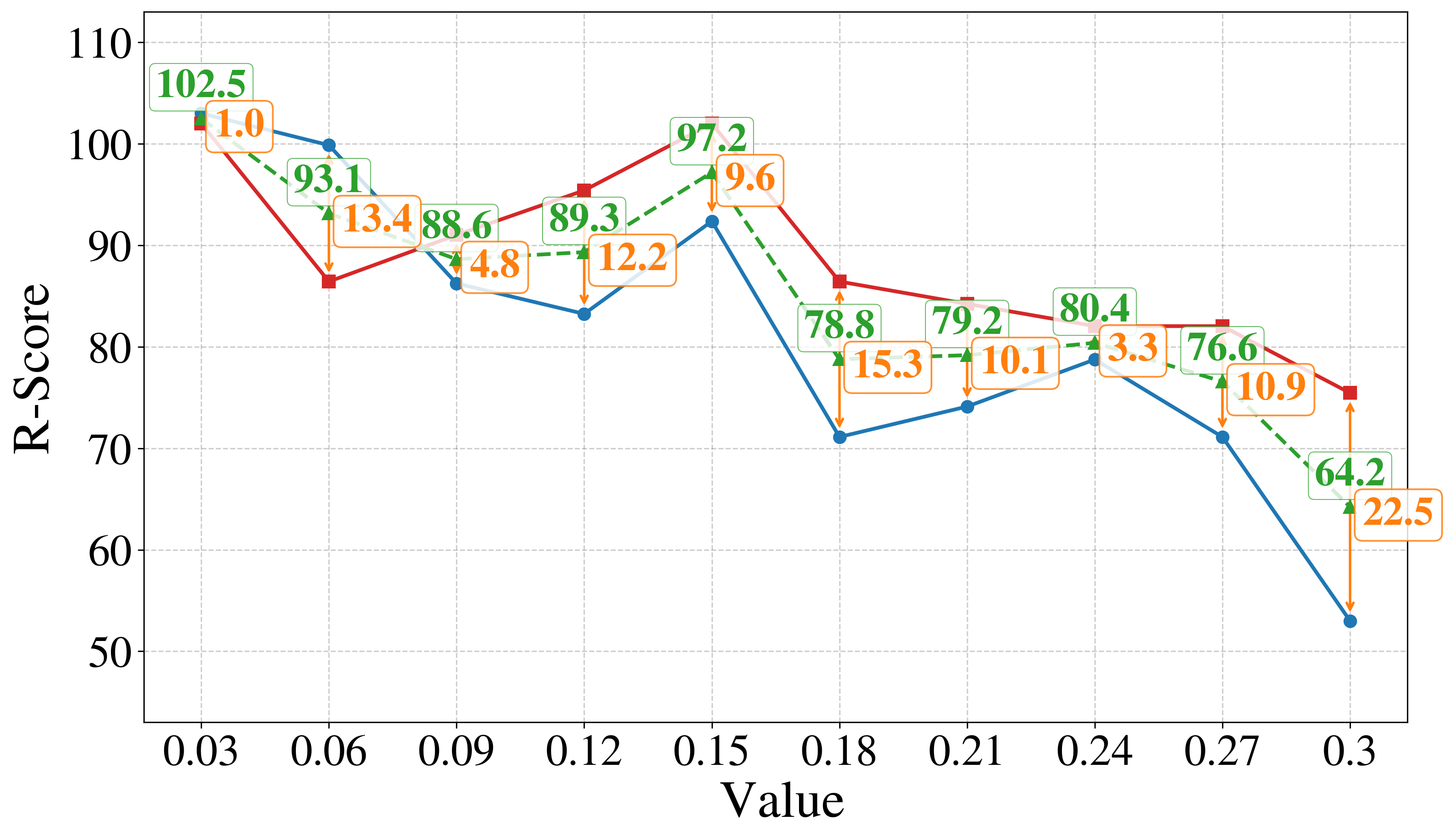}
    \par\vspace{-1mm}
    {\small\textbf{(c)}}\par
\end{minipage}%

\vspace{0.5mm}

\begin{minipage}[t]{0.33\textwidth}
    \centering
    \includegraphics[width=\linewidth]{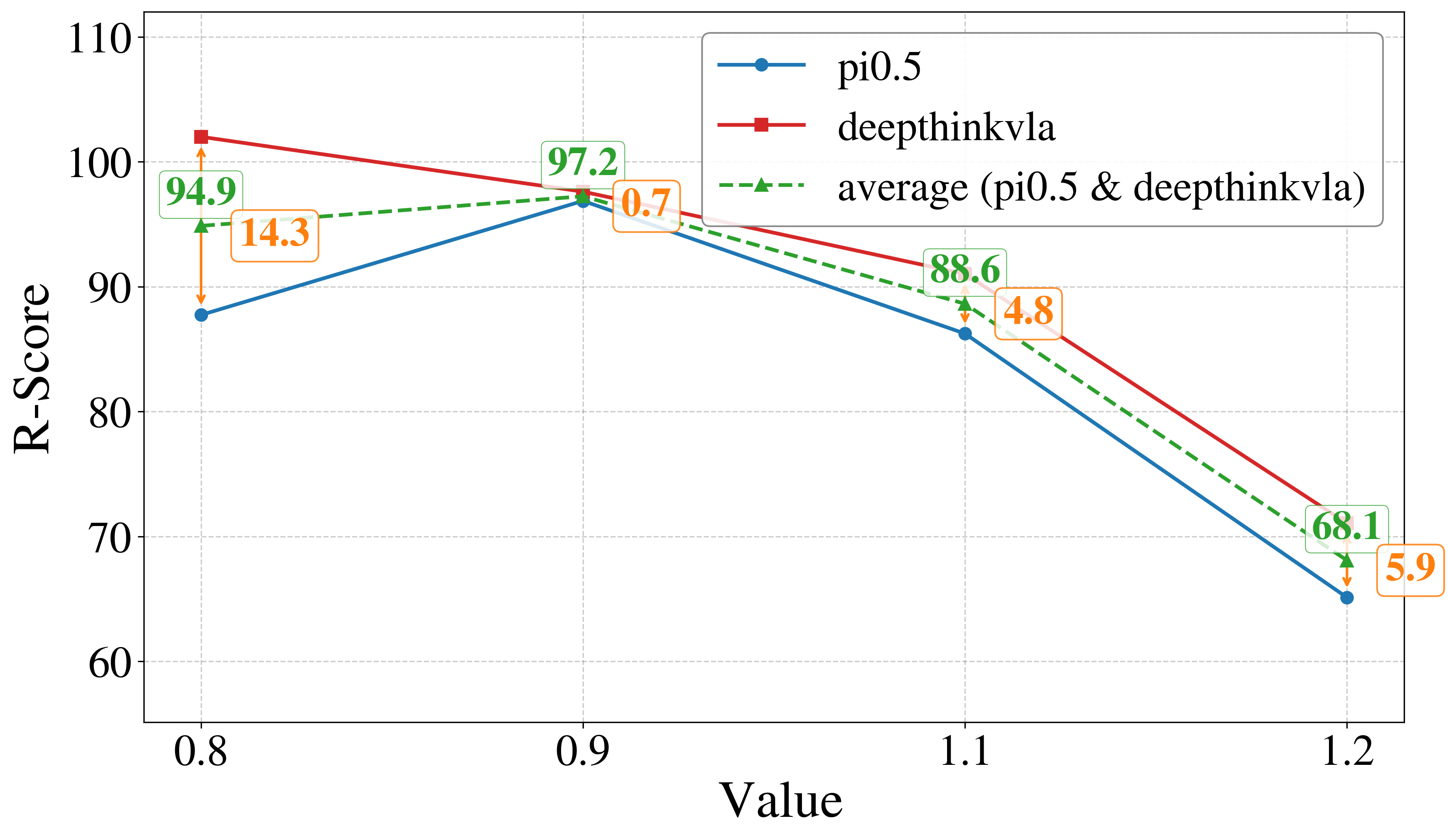}
    \par\vspace{-1mm}
    {\small\textbf{(d)}}\par
\end{minipage}%
\hfill
\begin{minipage}[t]{0.33\textwidth}
    \centering
    \includegraphics[width=\linewidth]{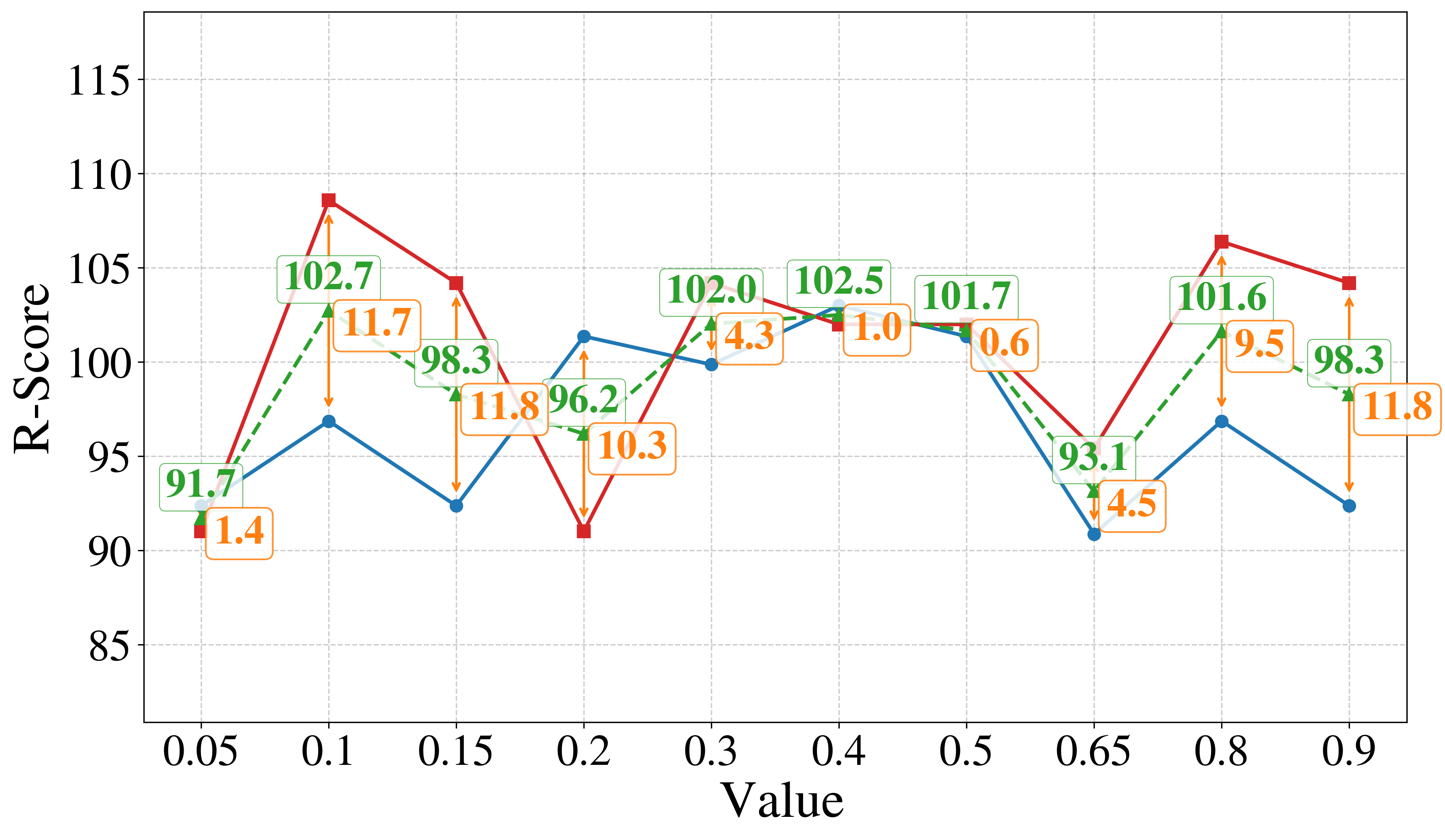}
    \par\vspace{-1mm}
    {\small\textbf{(e)}}\par
\end{minipage}%
\hfill
\begin{minipage}[t]{0.33\textwidth}
    \centering
    \includegraphics[width=\linewidth]{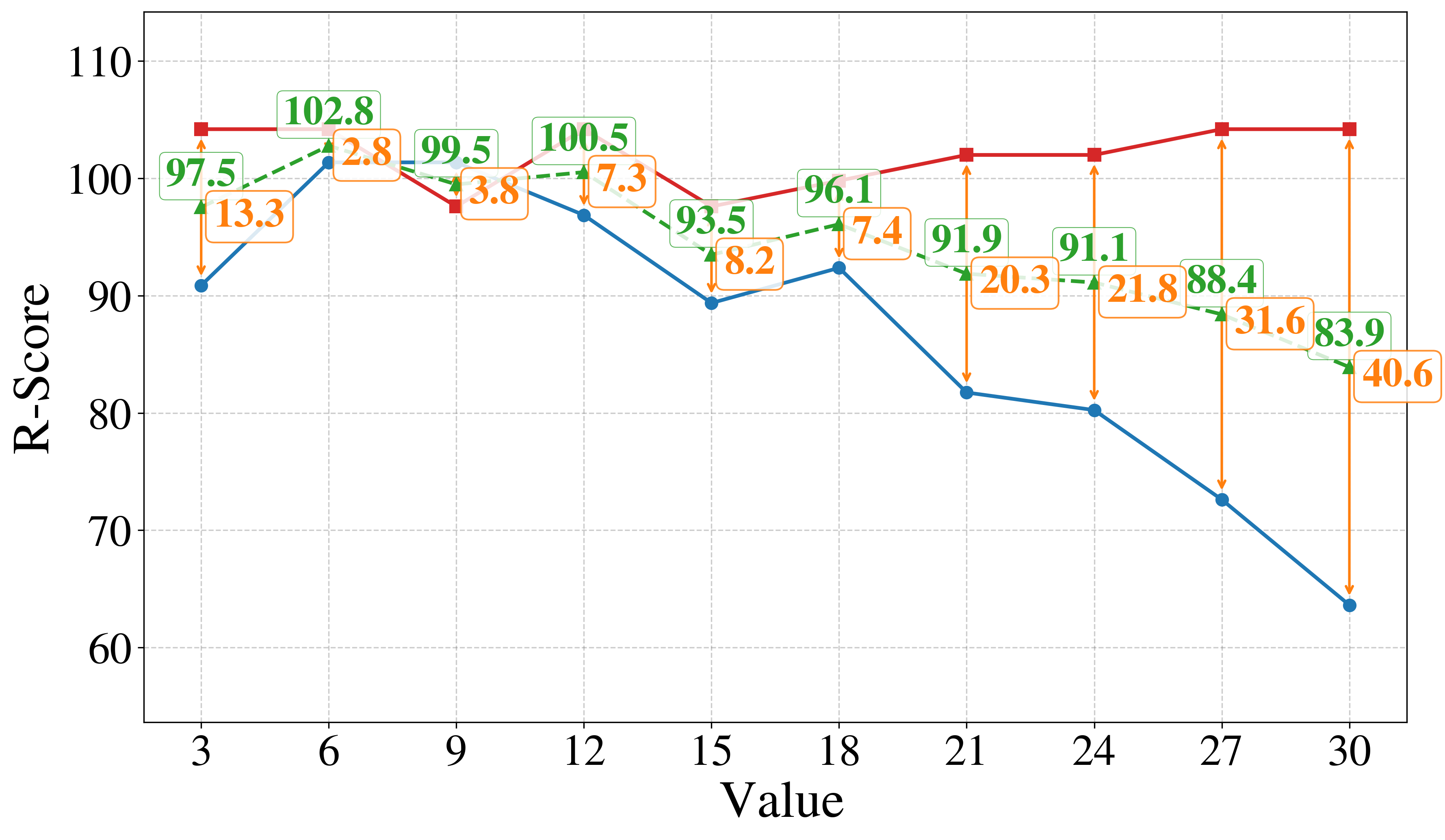}
    \par\vspace{-1mm}
    {\small\textbf{(f)}}\par
\end{minipage}%

\vspace{1mm}

\caption{\textbf{Sensitivity of the two reference models to six selected perturbation attributes.}
(a) Robot initial state.
(b) Action noise.
(c) Image shift.
(d) Target object scale.
(e) Light diffuse.
(f) Image rotation.}
\label{fig:single_factor_sensitivity}
\vspace{-1.2em}
\end{figure*}

\noindent\emph{\textbf{Results.}} We select 6 of the 33 perturbation attributes for focused analysis, as shown in Figure~\ref{fig:single_factor_sensitivity}. These attributes cover all three perturbation categories and clearly illustrate the magnitude of performance degradation, performance trends across parameter settings, and differences between the reference models.

As shown in Figure~\ref{fig:single_factor_sensitivity}(a) and~\ref{fig:single_factor_sensitivity}(b), perturbations to the robot's initial state and action noise markedly affect the performance of both reference models. For robot initial state, the mean $\mathrm{R\text{-}Score}$ across the two models ranges from \textcolor{blue}{13.6} to \textcolor{blue}{89.0} depending on the parameter setting. With a 475° offset in the robot's initial state, both models reach their respective minimum $\mathrm{R\text{-}Score}$. \emph{DeepThinkVLA}'s $\mathrm{R\text{-}Score}$ drops to zero, indicating greater sensitivity under this setting. For action noise, the mean $\mathrm{R\text{-}Score}$ ranges from \textcolor{blue}{30.4} to \textcolor{blue}{90.9} across parameter settings. As with initial state perturbations, some action noise settings cause substantial performance degradation.

\textbf{\emph{Finding~6:}~The sensitivity of the two reference models to changes in the robot initial state and to action noise varies markedly with the parameter setting, and some settings cause severe performance degradation.} These results suggest that developers deploying VLA models should pay particular attention to how the robot initial state is configured, and should limit the adverse effect of action noise on task execution.

For image shift, Figure~\ref{fig:single_factor_sensitivity}(c) shows that the mean R-Score across the two reference models decreases overall from \textcolor{blue}{102.5} to \textcolor{blue}{64.2} as the shift magnitude increases, despite occasional increases. For target object scale, Figure~\ref{fig:single_factor_sensitivity}(d) shows that both models maintain high $\mathrm{R\text{-}Score}$ at scale factors of 0.8 and 0.9. As the scale factor increases from 0.9 to 1.2, the mean $\mathrm{R\text{-}Score}$ decreases steadily, reaching a minimum of \textcolor{blue}{68.1}. One possible reason is that image shifts move some task-relevant information outside the model's field of view, leaving incomplete visual information, whereas enlarging the target object changes its apparent size and may make perception and grasping more difficult. In contrast, the mean $\mathrm{R\text{-}Score}$ for light diffuse remains between \textcolor{blue}{91.7} and \textcolor{blue}{102.7}, as shown in Figure~\ref{fig:single_factor_sensitivity}(e).

\textbf{\emph{Finding~7:}~Within the tested parameter ranges, the two reference models are more sensitive to large image shifts and to enlargement of the target object, whereas changes in diffuse illumination affect them relatively little.} Future work can explore strengthening contextual reasoning under incomplete visual observations and combining it with multi-scale visual augmentation, so as to improve the adaptability of a model to image shifts and to variations in target scale.

Figure~\ref{fig:single_factor_sensitivity}(f) shows that as the image rotation angle increases from 18° to 30°, \emph{$\pi$0.5}'s $\mathrm{R\text{-}Score}$ continues to decrease, whereas \emph{DeepThinkVLA}'s remains high with a slight overall increase. At 30°, the $\mathrm{R\text{-}Score}$ gap between the two models reaches \textcolor{blue}{40.6}, its maximum within the tested range.

\textbf{\emph{Finding~8:}~The two reference models can differ markedly in their sensitivity to the same image rotation setting.} Within the tested range of large rotation angles, \emph{$\pi$0.5} is considerably more sensitive to image rotation than \emph{DeepThinkVLA}. This suggests that model developers should pay close attention to how image rotation in the visual input affects model performance.

\vspace{-1.2em}
\subsection{\textcolor{green!75!black}{\faIcon{lightbulb}}~RQ4: Robustness to Combinatorial Perturbations}
\label{subsec:rq4}

\noindent\emph{\textbf{Methodology.}} To comprehensively evaluate the robustness of existing VLA models to combinatorial perturbations, we sequentially conducted combinatorial perturbation screening and robustness evaluation.

\emph{\textbf{Combinatorial Perturbation Screening}}. We follow the Stage~\ref{stage:2} rules in~\ref{subsec:perturbation_robustness_evaluation} to screen perturbation combinations. Using the~\ref{subsec:rq3} evaluation results, we identify sensitive value sets for each perturbation attribute and rank perturbation factors and their attributes by their impact on reference model performance. We first construct and evaluate 7 candidate perturbation category combinations (PCCs), then construct and evaluate 15 candidate perturbation factor combinations (PFCs) within the selected category combinations. 

In $\mathrm{PCC}(m,n_1)$ and $\mathrm{PFC}(m,n_2)$, $m$ denotes the number of perturbation categories used, $n_1$ and $n_2$ denote the category combination index and factor combination index, respectively, for that category count. For example, $\mathrm{PCC}(1,1)$ uses only visual information perturbations. $\mathrm{PFC}(2,2)$ combines spatial structure and control execution perturbations, instantiated through Target Object Scale, Action Noise, and Image Shift.

We use $\boldsymbol{\mathit{D\text{-}HM}}$ to assess whether a candidate combination causes pronounced performance degradation in all reference models. Both screening stages use the anchor scenario set $S_{anchor}$ and reference model set $M_{ref}$ and follow the same rule. After excluding candidates for which all reference models have an $\mathrm{R\text{-}Score}$ of zero, we select the combination with the highest $\mathrm{D\text{-}HM}$ separately among candidates involving one, two, or three perturbation categories. The selected representative PFCs are used for subsequent evaluation of robustness to combinatorial perturbations, with their configurations held fixed throughout evaluation.

\emph{\textbf{Combinatorial Perturbation Robustness Evaluation.}} Following the Stage~\ref{stage:3} evaluation protocol in~\ref{subsec:perturbation_robustness_evaluation}, we apply each of the three selected representative PFCs to all 270 unperturbed base scenarios, yielding 810 task instances with combinatorial perturbations. We evaluate all 12 VLA models on these instances. Using each model's unperturbed performance as the baseline, we use $\boldsymbol{\mathit{R\text{-}Score~(R)}}$ to measure its performance retention under combinatorial perturbations. We also use $\boldsymbol{\mathit{\%D\text{-}RMSD}}$ to quantify the overall performance degradation caused by each perturbation combination.

\begin{table}[t]
\centering
\caption{\textbf{Overall robustness under representative PFCs across VLA models.}}
\label{tab:pfc_summary}
\vspace{0.15em}

\begin{minipage}{0.98\columnwidth}
\centering
\scriptsize
\emph{
\textcolor{blue}{\textbf{Average1.}} and
\textcolor{blue}{\textbf{Average2.}}
denote per-model means across representative PFCs and column-wise
means across models, respectively.
\textcolor{blue}{\textbf{D-RMSD}} is computed from the original PFC
success rates relative to ZS. The \setlength{\fboxsep}{1pt}
\colorbox{bestbg}{\textcolor{blue}{top-1}} and
\colorbox{secondbg}{top-2}
results are highlighted, respectively. Near-zero $\mathrm{R\text{-}Scores}$ appear in \textcolor{red}{red}.
}
\end{minipage}

\vspace{0.25em}

\scriptsize
\setlength{\tabcolsep}{4.5pt}
\renewcommand{\arraystretch}{0.88}

\setlength{\aboverulesep}{0.22ex}
\setlength{\belowrulesep}{0.22ex}

\resizebox{\columnwidth}{!}{
\begin{tabular}{lccccc}
\toprule

\textbf{Models}
&
\emph{\textbf{Params}}
&
\emph{\textbf{PFC(1,2)}}
&
\emph{\textbf{PFC(2,3)}}
&
\emph{\textbf{PFC(3,6)}}
&
\textcolor{blue}{\emph{\textbf{Average1.}}}
\\

\midrule
\rowcolor{gray!10}
\multicolumn{6}{c}{\textbf{Large Models}}
\\
\midrule

UniVLA~\cite{bu2025univla}
& 9B
& 9.3
& 8.9
& \textcolor{red}{0.8}
& 6.4
\\

OpenVLA-OFT~\cite{kim2025fine}
& 7B
& 10.3
& 9.9
& 2.9
& 7.7
\\

RIPT-VLA~\cite{tan2025interactive}
& 7B
& 8.2
& 4.4
& 1.6
& 4.8
\\

SimpleVLA-RL~\cite{li2025simplevla}
& 7B
& 7.5
& 2.5
& 1.3
& 3.8
\\

CronusVLA~\cite{li2025cronusvla}
& 7B
& 17.3
& 3.2
& 1.8
& 7.4
\\

\midrule
\rowcolor{gray!10}
\multicolumn{6}{c}{\textbf{Small Models}}
\\
\midrule

$\pi$0.5~\cite{intelligence2025pi_}
& 3B
& \second{43.0}
& \best{17.7}
& \second{4.9}
& \best{21.9}
\\

Nora~\cite{hung2025nora}
& 3B
& 10.6
& 10.6
& 1.5
& 7.6
\\

GR00T1.5~\cite{nvidia2025gr00t}
& 3B
& 14.1
& \second{17.6}
& \best{7.7}
& 13.1
\\

DeepThinkVLA~\cite{yin2025deepthinkvla}
& 2.9B
& 12.8
& 13.4
& 3.9
& 10.0
\\

X-VLA~\cite{zheng2025x}
& 0.9B
& \best{52.4}
& 3.5
& \textcolor{red}{0.9}
& \second{18.9}
\\

Evo-1~\cite{lin2026evo}
& 0.77B
& 6.5
& 5.2
& 4.4
& 5.4
\\

VLA-Adapter-Pro~\cite{wang2026vla}
& 0.5B
& 6.4
& 4.8
& 3.2
& 4.8
\\

\midrule
\textcolor{blue}{\emph{\textbf{Average2.}}}
& --
& 16.5
& 8.5
& 2.9
& 9.3
\\

\midrule
\textcolor{blue}{\emph{\textbf{D-RMSD}}}
& --
& 19.3
& 22.7
& 24.8
& --
\\

\bottomrule
\end{tabular}
}

\vspace{-2.5em}
\end{table}

\begin{figure*}[t] \centering \vspace{-0.8em} \begin{minipage}[t]{0.494\textwidth} \centering \includegraphics[width=\linewidth]{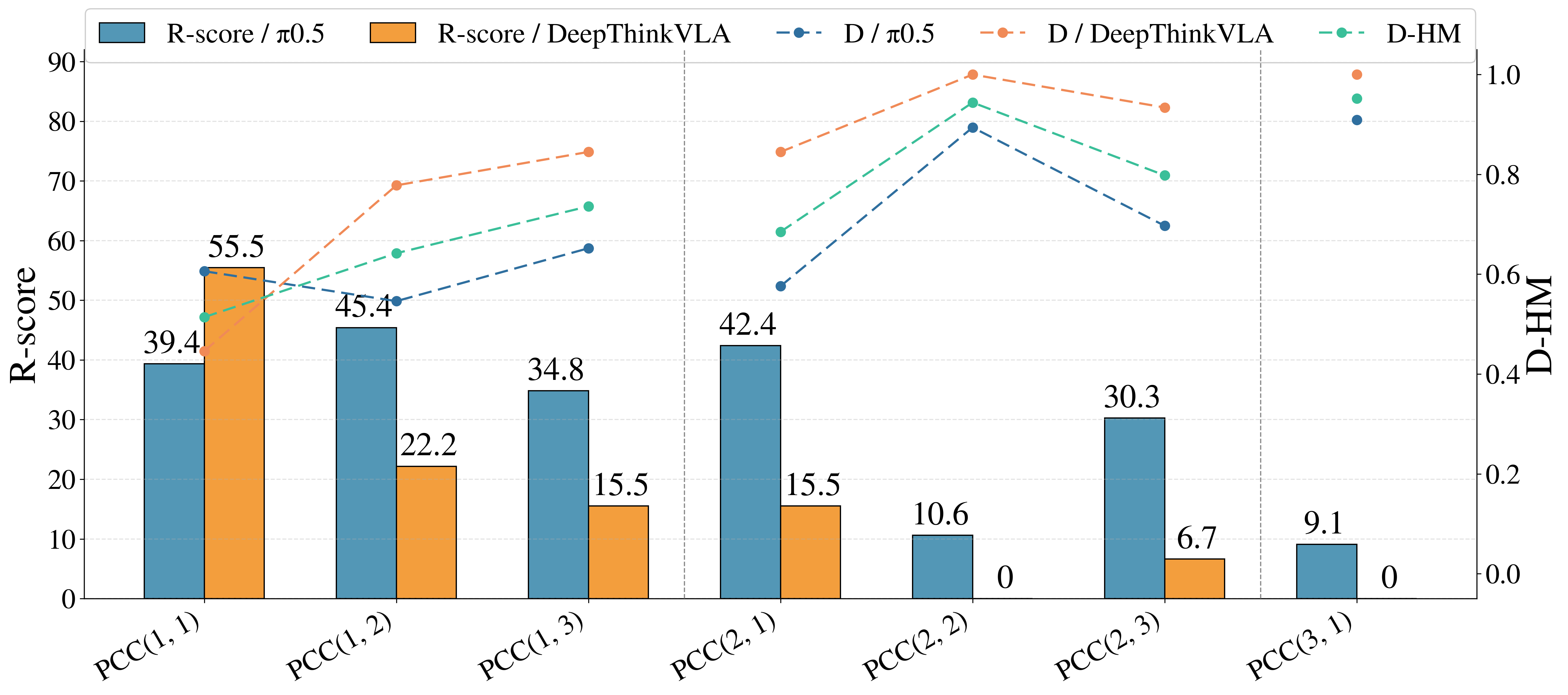} \vspace{-2mm} \makebox[\linewidth][c]{\small\textbf{(a)}} \end{minipage} \hfill \begin{minipage}[t]{0.494\textwidth} \centering \includegraphics[width=\linewidth]{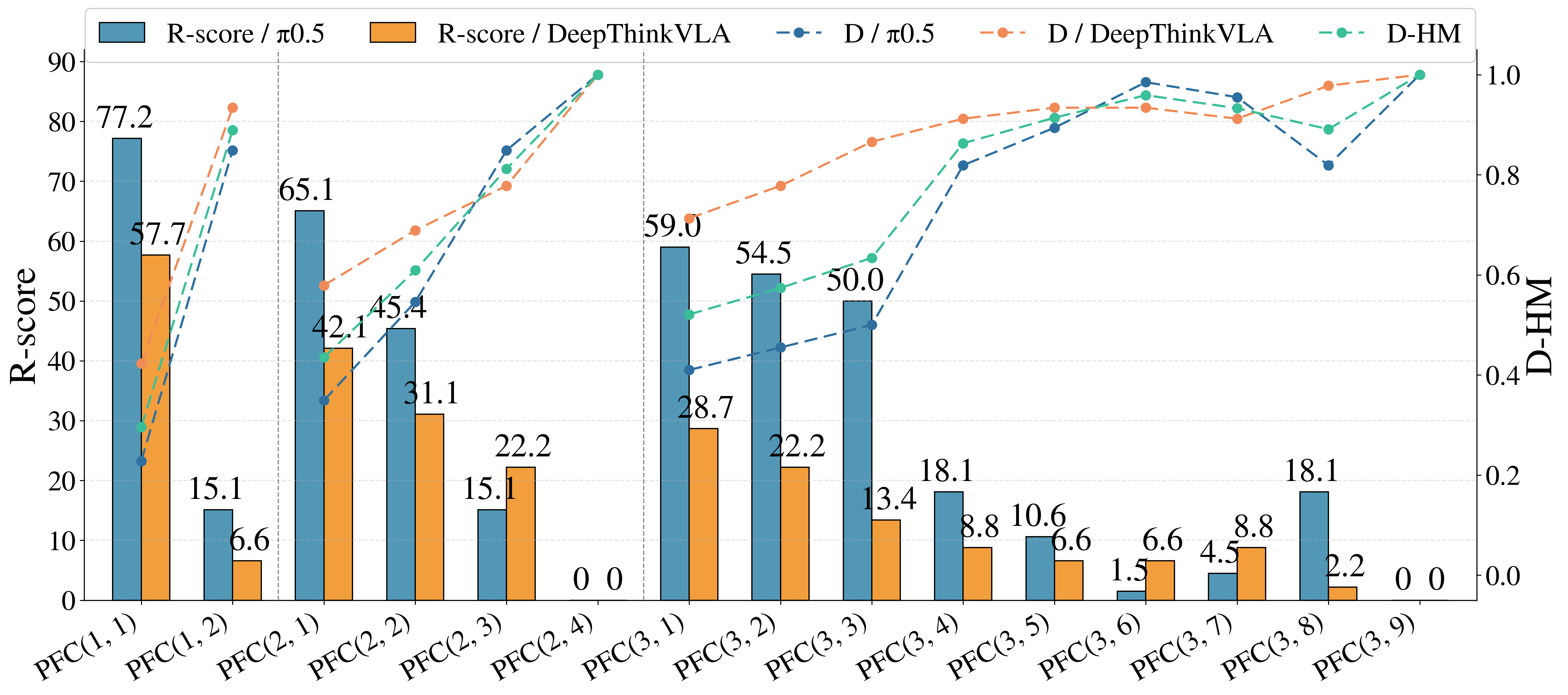} \vspace{-2mm} \makebox[\linewidth][c]{\small\textbf{(b)}} \end{minipage} \vspace{1mm} \caption{\textbf{Results of combinatorial perturbation selection.} (a) perturbation category combination selection. (b) perturbation factor combination selection.} \label{fig:combination_selection} \vspace{-1.5em} \end{figure*}

\noindent\emph{\textbf{Results.}} As shown in Figure~\ref{fig:combination_selection}(a), $\mathrm{PCC}(1,3)$ and $\mathrm{PCC}(2,2)$ have the highest $\mathrm{D\text{-}HM}$ among candidates using one and two perturbation categories, respectively. $\mathrm{PCC}(3,1)$ is the only candidate using three categories. None of these combinations yields zero $\mathrm{R\text{-}Score}$ for both reference models, so all three are retained under the screening rules. Under $\mathrm{PCC}(1,3)$, which contains only control execution perturbations, \emph{$\pi$0.5} and \emph{DeepThinkVLA} achieve $\mathrm{R\text{-}Score}$ of only \textcolor{blue}{34.8} and \textcolor{blue}{15.5}, respectively. Both models have even lower $\mathrm{R\text{-}Score}$ under $\mathrm{PCC}(2,2)$ and $\mathrm{PCC}(3,1)$. These results show that both reference models struggle to maintain their unperturbed performance under these representative PCCs in the anchor scenarios.

Figure~\ref{fig:combination_selection}(b) further presents the evaluation results for the PFC candidates within the three representative PCCs. $\mathrm{PFC}(2,4)$ and $\mathrm{PFC}(3,9)$ are excluded because each yields zero $\mathrm{R\text{-}Score}$ for both reference models. Among the remaining candidates, $\mathrm{PFC}(1,2)$, $\mathrm{PFC}(2,3)$, and $\mathrm{PFC}(3,6)$ have the highest $\mathrm{D\text{-}HM}$ for combinations involving one, two, and three perturbation categories, respectively. We retain these three PFCs for subsequent evaluation of robustness to combinatorial perturbations.

$\mathrm{PFC}(1,2)$ includes Action noise and Robot initial state; $\mathrm{PFC}(2,3)$ includes Target object scale, action noise, image shift, and camera Vertical; $\mathrm{PFC}(3,6)$ includes Target object scale, background, action noise, image occlusions, light direction, image shift, placed object color, and noise. Both reference models have low $\mathrm{R\text{-}Score}$ under all three selected PFCs, indicating that these combinations cause marked performance degradation in both models.

As shown in Table~\ref{tab:pfc_summary}, the evaluated models achieve an overall mean $\mathrm{R\text{-}Score}$ of only \textcolor{blue}{9.3}. In the 36 evaluation results, the two highest $\mathrm{R\text{-}Scores}$ are \textcolor{blue}{52.4} and \textcolor{blue}{43.0}, while the remaining \textcolor{blue}{34} are not higher than \textcolor{blue}{17.7}. Across $\mathrm{PFC}(1,2)$, $\mathrm{PFC}(2,3)$, and $\mathrm{PFC}(3,6)$, the mean $\mathrm{R\text{-}Score}$ decreases and $\mathrm{D\text{-}RMSD}$ increases at each step.

\textbf{\emph{Finding~9:}~ The evaluated VLA models exhibit insufficient overall robustness under the three representative combinatorial perturbations, and the overall performance loss becomes progressively more severe as the selected combinations move from a single perturbation category to three categories.} \textsc{VLA-Gauge} organizes its representative combinations by the number of perturbation categories involved, which makes it possible to assess the joint effect of multiple perturbation types under controlled conditions and thereby provides a basis for identifying potential performance problems under combinatorial perturbations before real deployment.

We also find that model robustness rankings vary across representative perturbation combinations. \emph{X-VLA} achieves the highest $\mathrm{R\text{-}Score}$ under $\mathrm{PFC}(1,2)$, but scores only \textcolor{blue}{3.5} and \textcolor{blue}{0.9} under $\mathrm{PFC}(2,3)$ and $\mathrm{PFC}(3,6)$, respectively. In contrast, \emph{GR00T1.5} achieves an $\mathrm{R\text{-}Score}$ close to that of the leading model \emph{$\pi$0.5} under $\mathrm{PFC}(2,3)$ and ranks first under $\mathrm{PFC}(3,6)$ with an $\mathrm{R\text{-}Score}$ of \textcolor{blue}{7.7}.

\textbf{\emph{Finding~10:}~None of the evaluated models ranks first under all three groups of representative PFCs.} Therefore, assessing the combinatorial perturbation robustness of a VLA model requires considering its performance across different compositions jointly.

\begin{figure}[htbp]
\centering
\vspace{-0.6em}
\includegraphics[width=\linewidth]{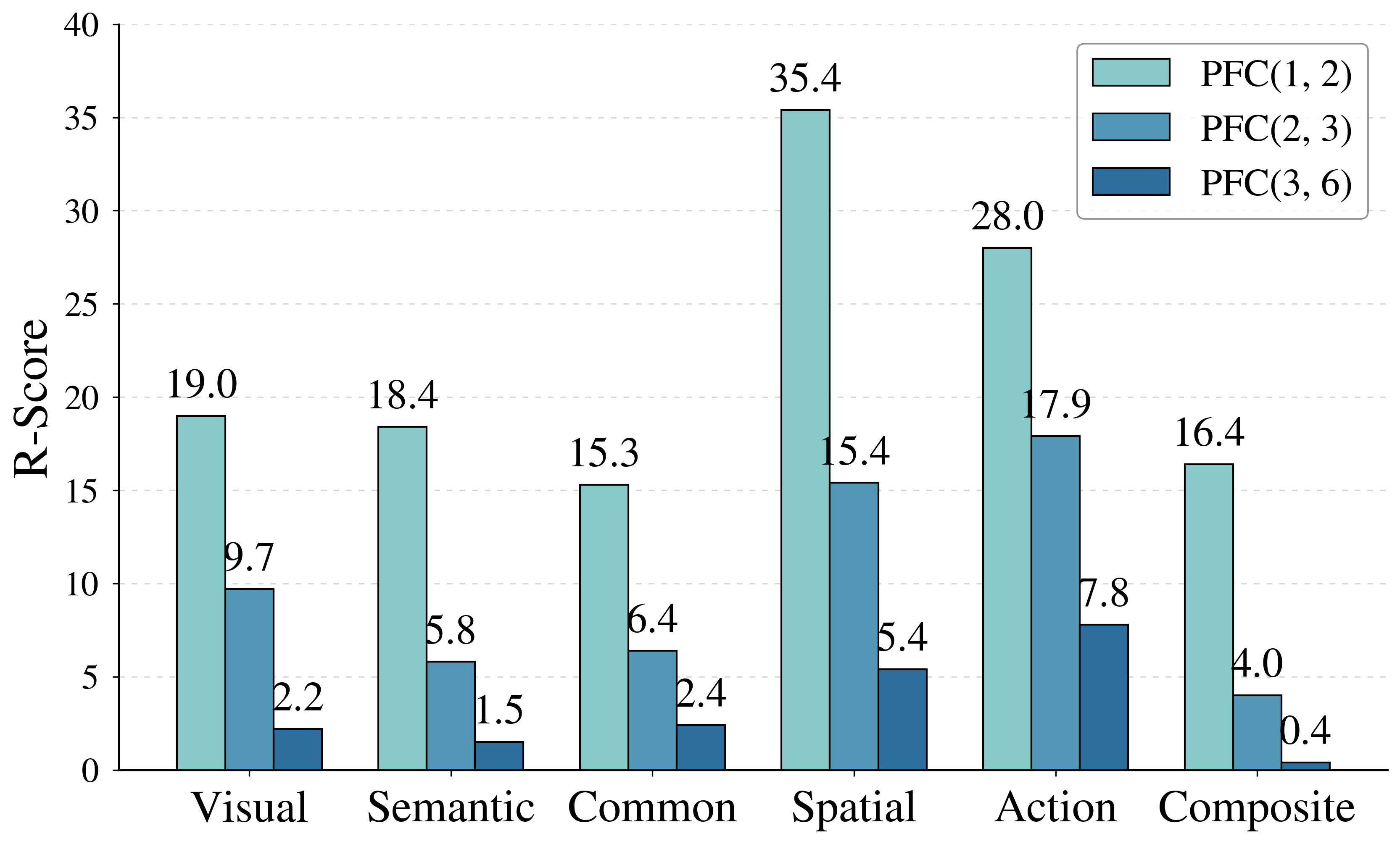}
\caption{\textbf{Capability robustness under representative combinatorial perturbations.}}
\label{fig:capability_average_robustness}
\vspace{-1.2em}
\end{figure}

Figure~\ref{fig:capability_average_robustness} shows that the mean $\mathrm{R\text{-}Scores}$ across the 12 VLA models are generally low across the six capability dimensions under the three PFCs. Spatial reasoning and action reasoning have higher mean $\mathrm{R\text{-}Score}$ than the other four dimensions under all three PFCs, whereas compositional capability has the lowest mean $\mathrm{R\text{-}Score}$ under $\mathrm{PFC}(2,3)$ and $\mathrm{PFC}(3,6)$. One possible explanation is that the evaluated models' training data do not adequately cover diverse combinations of perturbation settings. Future work may use multiple parameter settings for each factor in a perturbation combination to construct training examples covering diverse perturbation states. This strategy aims to improve robustness to combinatorial perturbations across all six capability dimensions, with particular emphasis on compositional capability.

\textbf{\emph{Finding~11:}~All six capability dimensions of the evaluated models degrade under combinatorial perturbations.} Relatively strong performance on an individual capability does not imply comparably strong combinatorial perturbation robustness on compositional tasks that require several capabilities to be exercised together.

\vspace{-1.2em}
\subsection{\textcolor{green!75!black}{\faIcon{lightbulb}}~RQ5: Effectiveness of diagnosis-driven Fine-Tuning}
\label{subsec:rq5}

\noindent\emph{\textbf{Methodology.}}~We follow the diagnosis-driven optimization protocol in~\ref{subsec:diagnostic_guided_model_refinement} to test whether \textsc{VLA-Gauge's} fine-grained diagnostics help construct effective training data. We select $\pi_{0.5}$ as the target model because it achieves the highest average performance in the zero-shot generalization and combinatorial perturbation robustness evaluations. Based on the results in~\ref{subsec:rq4}, we select $\mathrm{PFC}(2,3)$, which has the intermediate $\mathrm{D\text{-}RMSD}$ value among the three representative PFCs, to optimize and evaluate robustness to combinatorial perturbations. 

In the zero-shot generalization branch, we construct training data following the augmentation rules in~\ref{subsec:diagnostic_guided_model_refinement}, guided by the diagnostic results and optimization recommendations from~\ref{subsec:rq1} and~\ref{subsec:rq2}. The specific settings are as follows:

\begin{itemize}[leftmargin=*, itemsep=0.2em, topsep=0.2em]
    \item To address the shortfall in spatial reasoning performance under ID conditions, we start from 4 challenging ID training scenarios and sample translation distances uniformly from 1 to 5 cm. We generate 8 new variants per original scenario, yielding 32 additional training scenarios in total.
    \item To address the performance shortfall in compositional capabilities under OOD conditions, we start from 5 challenging OOD training scenarios. For each original scenario, we construct compositional tasks involving 2, 3, and 4 basic capabilities, adding one new scenario per category and yielding 15 additional training scenarios in total.
    \item To address the overall shortfall in zero-shot generalization across the remaining four capability dimensions, we augment the 12 existing challenging training scenarios. For semantic grounding, visual perception, and common sense and world knowledge, we generate one variant per original scenario using each of the three augmentation methods defined for the corresponding capability. For action reasoning, we add 3 scenarios per original scenario, each using a different instance of the manipulated object. This yields 12 additional scenarios from the 4 semantic grounding scenarios and 24 additional scenarios from the 8 scenarios covering the other three capabilities.
\end{itemize}

In the combinatorial perturbation robustness branch, we address the insufficient robustness of the evaluated models across all six capability dimensions revealed by~\ref{subsec:rq4}. Taking the $\mathrm{PFC}(2,3)$ parameter configuration retained during screening as the baseline, we follow the augmentation rules in~\ref{subsec:diagnostic_guided_model_refinement} and randomly sample, from the candidate set of each of its four perturbation attributes, one replacement value distinct from the original. Replacing one attribute value at a time thus yields four new combinatorial perturbation states. Applying each of these four states to the 45 existing challenging training scenarios yields 180 new training scenarios, giving 225 scenarios in total together with the original ones.

Both optimization branches follow the collection and quality verification procedure described in~\ref{subsec:diagnostic_guided_model_refinement}, recording multi-view visual observations, end-effector states, and action sequences synchronously at 20 Hz. Each training scenario contributes 50 successful trajectories, so the two types of optimization data together comprise 16,450 successful trajectories. We compare the zero-shot generalization score and the combinatorial perturbation robustness score under $\mathrm{PFC}(2,3)$ before and after fine-tuning $\pi_{0.5}$. For the robustness branch, we additionally report fine-tuning on the single-factor perturbation data of \textsc{Libero-Plus} as a reference, a dataset that contains more successful trajectories than the training set of our robustness branch.

\begin{figure*}[t]
\centering

\begin{minipage}[t]{0.185\textwidth}
    \vspace{0pt}
    \centering
    \includegraphics[width=\linewidth]{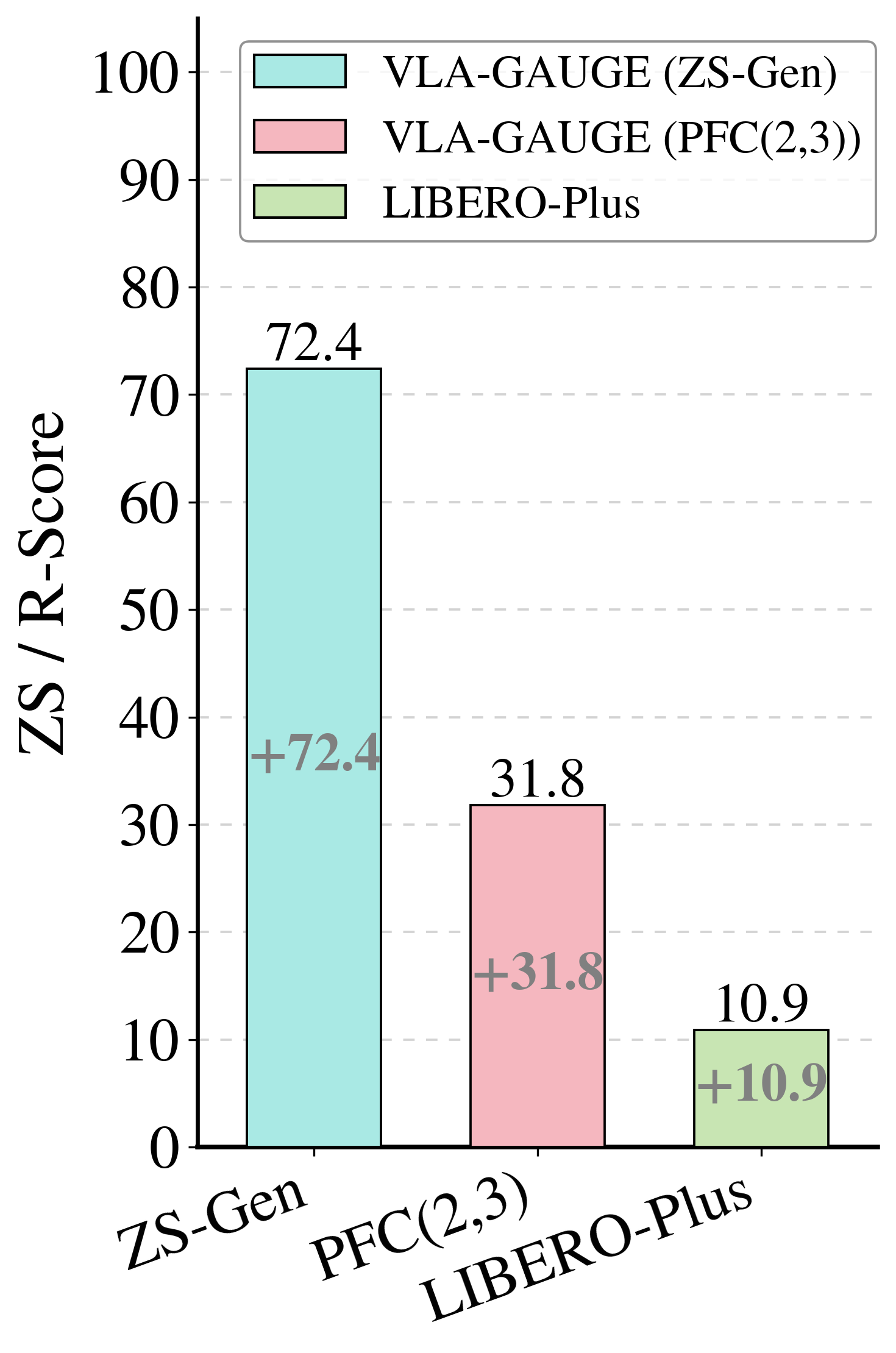}
    \par\vspace{0.1em}\textbf{(a)}
\end{minipage}\hspace{0.005\textwidth}%
\begin{minipage}[t]{0.375\textwidth}
    \vspace{0pt}
    \centering
    \includegraphics[width=\linewidth]{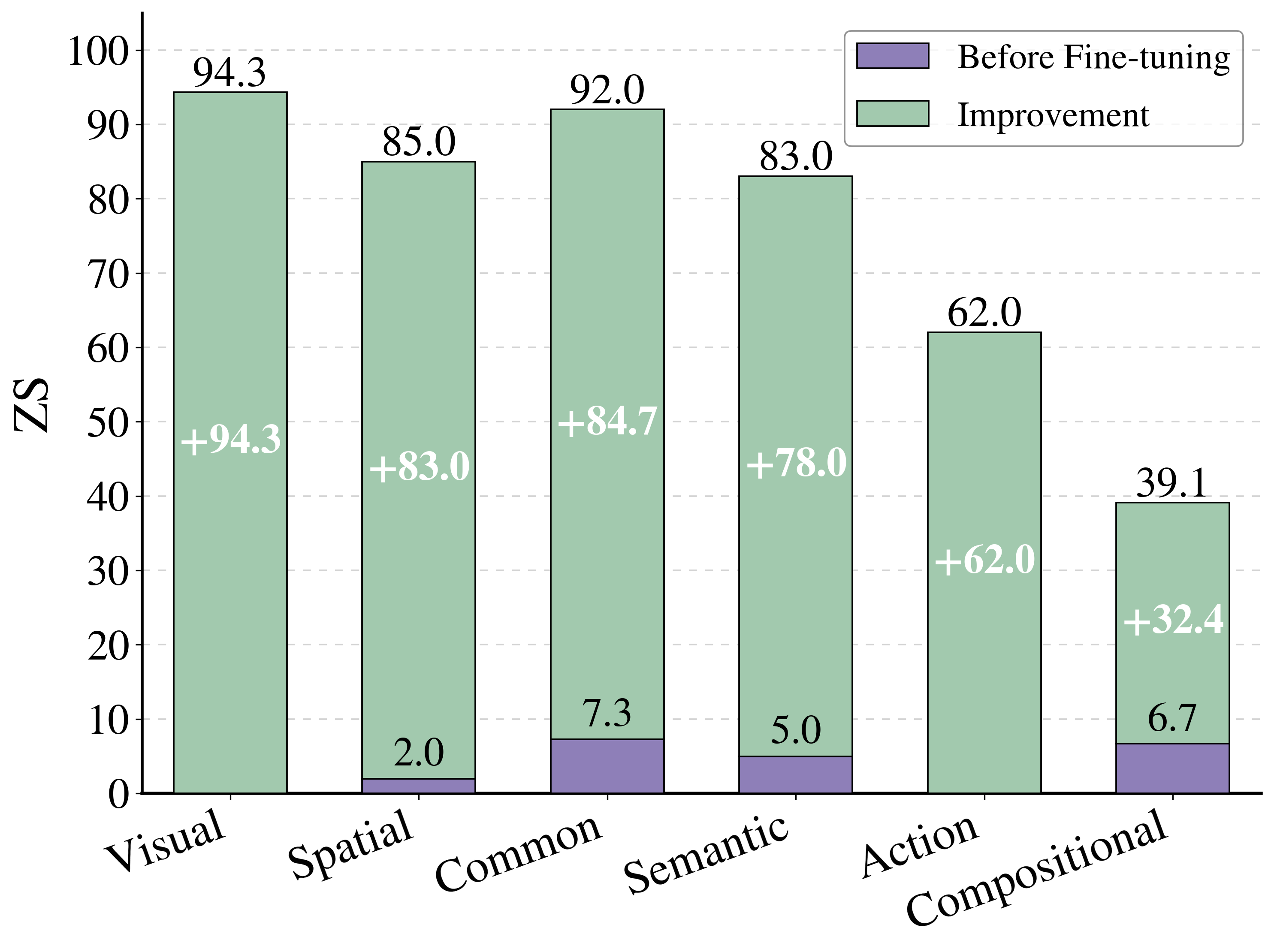}
    \par\vspace{0.1em}\textbf{(b)}
\end{minipage}\hspace{0.005\textwidth}%
\begin{minipage}[t]{0.375\textwidth}
    \vspace{0pt}
    \centering
    \includegraphics[width=\linewidth]{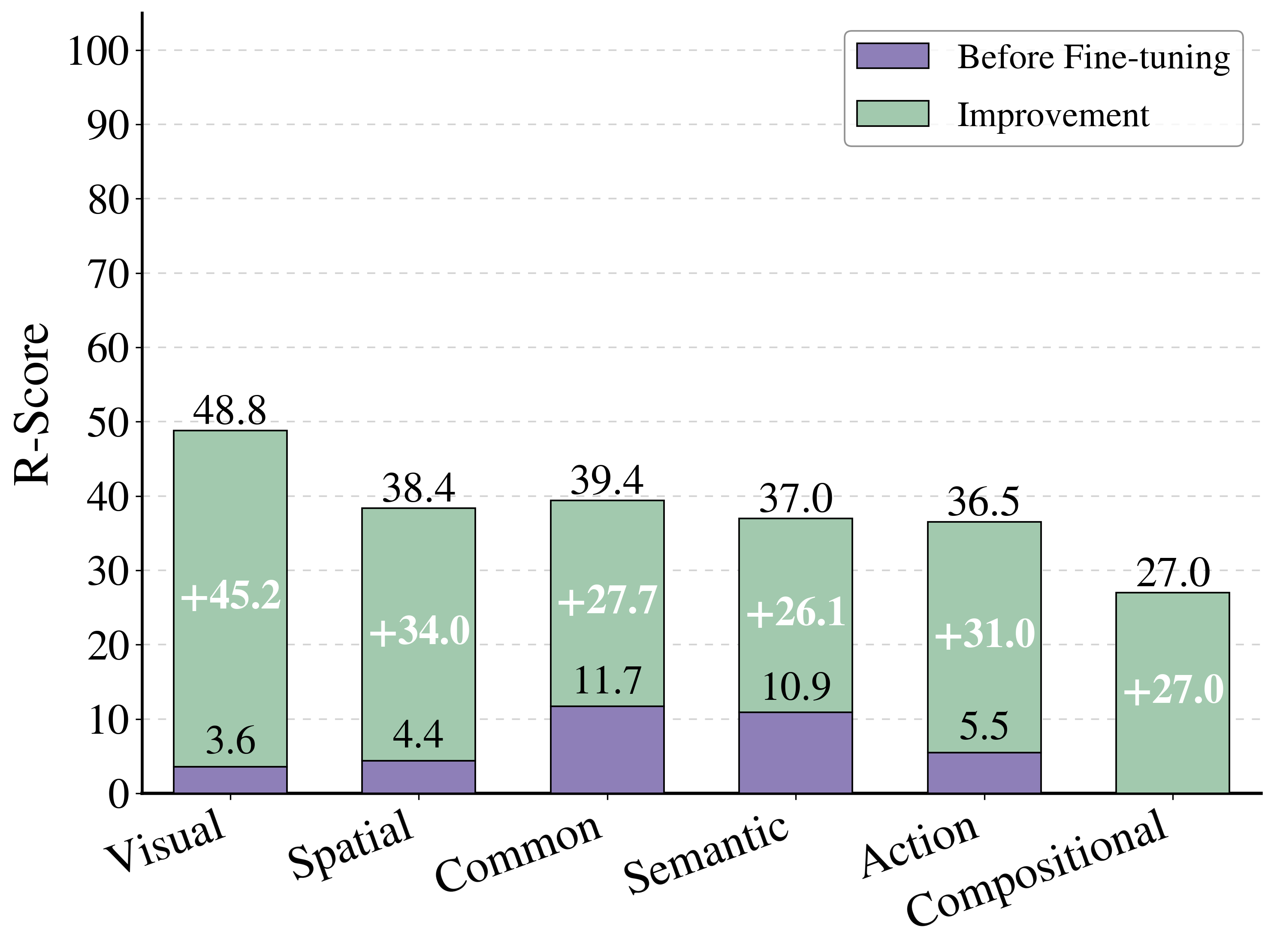}
    \par\vspace{0.1em}\textbf{(c)}
\end{minipage}

\caption{\textbf{Diagnosis-driven fine-tuning gains in zero-shot generalization and perturbation robustness.}
(a) Overall comparison of the fine-tuning gains of \textsc{VLA-Gauge} on zero-shot generalization and on $\mathrm{PFC}(2,3)$, together with the robustness gain from fine-tuning on \textsc{Libero-Plus} single-factor perturbation data. (b) Zero-shot generalization gains across capabilities. (c) Robustness gains under PFC(2,3) across capabilities.} 
\label{fig:finetuning_gain_comparison}
\vspace{-1.5em}
\end{figure*}

\noindent\emph{\textbf{Results.}} As shown in Figure~\ref{fig:finetuning_gain_comparison}(a), our method increases the target model's mean zero-shot generalization score by \textcolor{blue}{$\Uparrow$72.4} points and its mean robustness score under $\mathrm{PFC}(2,3)$ by \textcolor{blue}{$\Uparrow$31.8} points. In comparison, fine-tuning on \textsc{Libero-Plus} single-factor perturbation data improves the mean robustness score by only \textcolor{blue}{$\Uparrow$10.9} points.

\textbf{\emph{Finding~12:}~The optimization data constructed under the diagnosis-driven procedure of \textsc{VLA-Gauge} effectively improves both the zero-shot generalization of the target model and its combinatorial perturbation robustness under $\boldsymbol{\mathrm{PFC}(2,3)}$.} Notably, compared with the single-factor fine-tuning baseline, our approach achieves a larger robustness gain under the $\mathrm{PFC}(2,3)$ combinatorial perturbation condition while using fewer successful training trajectories.

Figure~\ref{fig:finetuning_gain_comparison}(b) and Figure~\ref{fig:finetuning_gain_comparison}(c) show marked score gains across all six capability dimensions in both branches. In the zero-shot generalization branch, spatial reasoning, compositional capability, and semantic grounding are key optimization targets, with score gains of \textcolor{blue}{$\Uparrow$83.0}, \textcolor{blue}{$\Uparrow$32.4}, and \textcolor{blue}{$\Uparrow$78.0} points, respectively. The $\mathrm{PFC}(2,3)$ robustness branch achieves gains of \textcolor{blue}{$\Uparrow$26.1} to \textcolor{blue}{$\Uparrow$45.2} points across the six dimensions.

\textbf{\emph{Finding~13:}~The optimization data constructed under the diagnosis-driven procedure of \textsc{VLA-Gauge} improves the capabilities on which the target model performs relatively weakly, in both the zero-shot generalization branch and the $\boldsymbol{\mathrm{PFC}(2,3)}$ robustness branch, which demonstrates the practical value of fine-grained diagnosis for guiding model optimization.}

\section{Discussions}

\emph{\textbf{Data Leakage:}} Because the full pretraining data of the 12 evaluated VLA models cannot be thoroughly verified, we cannot entirely rule out the possibility that their pretraining content overlaps with the evaluation tasks in \textsc{VLA-Gauge}. To reduce the risk of data leakage in our evaluation procedure, we use the officially released model checkpoints without additional training or adaptation and keep all model parameters frozen during the initial evaluations. In the diagnosis-driven fine-tuning stage, we split the challenging scenarios of each of the two evaluation branches into training and test sets that share no tasks or scene configurations, and we perform data augmentation and construct the optimization data using only the training scenarios.

\emph{\textbf{Threats to Validity:}} One potential threat is that \textsc{VLA-Gauge} is built on the \textsc{Libero} simulation environment, so its evaluation scope may not fully cover other robot embodiments or the diverse conditions of real-world deployment. To broaden the coverage of the simulated evaluation, we construct a hierarchical perturbation space that spans three types of perturbations, namely visual information, spatial structure, and control execution perturbations, and we evaluate 12 VLA models on 270 base scenarios covering six capability dimensions, together with their perturbed instances. Future work can extend the evaluation to other robot embodiments and to real-robot tasks, which would test whether the capability gaps and perturbation sensitivities observed here still hold under these conditions.

\section{Conclusion}

We present \textsc{VLA-Gauge}, the first unified diagnostic framework for systematically assessing the zero-shot generalization and perturbation robustness of VLA models. The framework integrates six fine-grained capability suites with paired in-distribution (ID) and out-of-distribution (OOD) splits and a hierarchical perturbation space, and it closes the optimization loop through diagnosis-driven data construction and fine-tuning. Our large-scale empirical study of 12 VLA models shows that high reported in-distribution success rates do not guarantee strong zero-shot generalization to unseen tasks or robustness to combinatorial perturbations. The diagnosis-driven fine-tuning experiments further show that fine-grained diagnostic results, together with the optimization suggestions derived from them, effectively guide the construction of training data and improve both the zero-shot generalization and combinatorial perturbation robustness of the target model. We therefore urge the community to give these two capabilities greater weight in the evaluation and development of VLA models.

\bibliographystyle{IEEEtran}
\bibliography{software}

\end{document}